\documentclass[journal]{vgtc}                     

\onlineid{1789}

\vgtccategory{Research}

\vgtcpapertype{system}

\title{Int3DNet: Scene–Motion Cross Attention Network \\ for 3D Intention Prediction in Mixed Reality}

\author{%
  \authororcid{Taewook Ha}{0009-0004-7946-8577},
  \authororcid{Woojin Cho}{0009-0003-4615-5630},
  \authororcid{Dooyoung Kim}{0000-0002-6003-2181}, and
  \authororcid{Woontack Woo}{0000-0002-5501-4421}
}

\authorfooter{
  \item
  	Taewook Ha is with KAIST UVR Lab. E-mail: tw.ha@kaist.ac.kr
  \item
  	Woojin Cho is with KAIST KI-ITC PMRC. E-mail: woojin.cho@kaist.ac.kr
  \item
  	Dooyoung Kim is with KAIST KI-ITC ARRC. E-mail: dooyoung.kim@kaist.ac.kr
  \item
  	Woontack Woo is with KAIST UVR Lab and KAIST KI-ITC ARRC. E-mail: wwoo@kaist.ac.kr
}

\abstract{
  We propose Int3DNet, a scene-aware network that predicts 3D intention areas directly from scene geometry and head–hand motion cues, enabling robust human intention prediction without explicit object-level perception. In Mixed Reality (MR), intention prediction is critical as it enables the system to anticipate user actions and respond proactively, reducing interaction delays and ensuring seamless user experiences. Our method employs a cross attention fusion of sparse motion cues and scene point clouds, offering a novel approach that directly interprets the user's spatial intention within the scene. We evaluated Int3DNet on MoGaze and CIRCLE datasets, which are public datasets for full-body human–scene interactions, showing consistent performance across time horizons of up to 1500 ms and outperforming the baselines, even in diverse and unseen scenes. Moreover, we demonstrate the usability of proposed method through a demonstration of efficient visual question answering (VQA) based on intention areas. Int3DNet provides reliable 3D intention areas derived from head–hand motion and scene geometry, thus enabling seamless interaction between humans and MR systems through proactive processing of intention areas.
}


\teaser{
  \centering
  \includegraphics[width=\columnwidth]{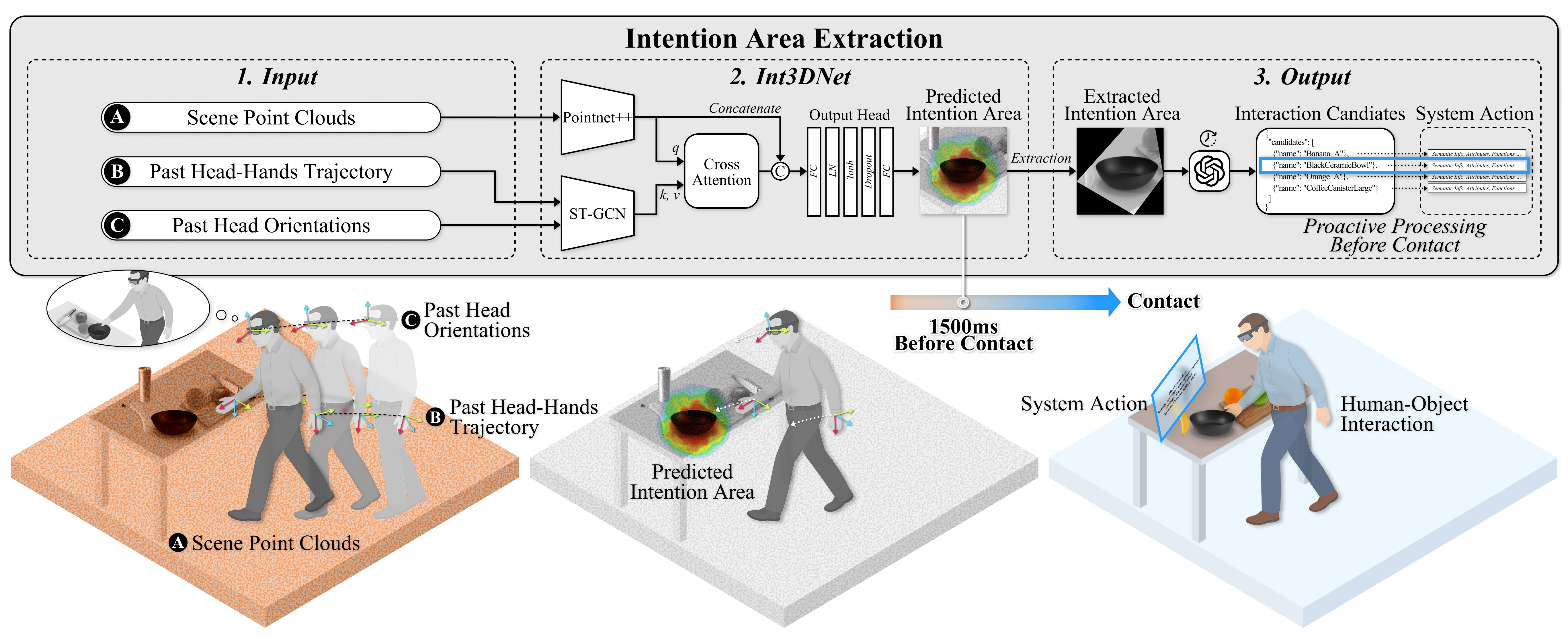}
  \caption{%
  Given scene point clouds, past head–hand trajectories, and head orientations, Int3DNet predicts the future intention area within the 3D scene. Leveraging this prediction, possible applications such as intention-based VQA can be achieved by feeding the extracted intention area into a VLM, enabling proactive processing before contact.
  }
  \label{fig:teaser}
}

\graphicspath{{figs/}{figures/}{pictures/}{images/}{./}} 

\usepackage{tabu}                      
\usepackage{booktabs}                  
\usepackage{lipsum}                    
\usepackage{mwe}                       

\usepackage{mathptmx}                  
\usepackage{amssymb}
\usepackage{amsmath}
\usepackage{amsthm}
\usepackage{amsfonts}
\usepackage{amstext}
\usepackage{multirow}
\usepackage{enumitem}

\begin{document}


\firstsection{Introduction}

\maketitle

Understanding human intent to interact is a crucial factor in the interaction between humans and systems, whether they are 2D user interfaces~\cite{koochaki2019eye,  chang2021temporal}, embodied robots~\cite{mainprice2013human, choi2022preemptive, laplaza2024enhancing}, or immersive systems~\cite{clarence2021unscripted, david2021towards, gamage2021so, du2025predicting}. Intention recognition enables systems to adaptively align their behavior with user goals and facilitates more natural collaboration. Crucially, anticipating user intentions in advance allows systems to deliver proactive and seamless feedback with minimal latency. For instance, if reasoning begins only after an interaction, delays are unavoidable; by contrast, proactive intention inference enables reasoning beforehand, ensuring information is available precisely at the moment of interaction.

The time gap between interaction and reasoning is particularly critical in Mixed Reality (MR) systems, where even minor delays can significantly disrupt user immersion~\cite{caserman2019effects, brunnstrom2020latency}. To address this, few studies have been conducted to construct sensory feedback systems~\cite{gamage2021so, clarence2021unscripted} or propose predictive interfaces~\cite{david2021towards} by preemptively understanding human intention. In addition, recent work has focused on understanding future user intention by forecasting a user's hand trajectory from an egocentric view~\cite{bao2023uncertainty}. While prior studies have focused on inferring intention from user motion alone, research on interpreting intention while considering the physical environment in MR systems remains relatively limited. In contrast, the robotics and human-robot interaction (HRI) fields have actively investigated this problem, particularly in the context of collaborative and manipulation tasks. For instance, previous studies~\cite{leon2022intuitive, kratzer2020anticipating} proposed an intention prediction system within a 3D scene to avoid collisions with collaborative robots while ensuring immediate and responsive collaboration.  
Inspired by the similarity that both MR systems and embodied systems must consider human interaction within the physical environment, we aim to adapt the 3D scene-aware intention prediction paradigm from the HRI field to the context of MR systems. 

\newpage

Despite the similarity between MR and embodied systems, intention prediction methods from embodied settings require adaptation for MR. In particular, intention in MR is shaped by the complex correlation of multiple cues, including motion, gaze, and spatial context, and the key challenge is how to effectively capture these cues within MR settings. Specifically, prior research~\cite{hoffman2023inferring, shi2021understand, kedia2024interact, laplaza2022context, wei2017inferring} could obtain full-body or upper-limb motion cues leveraging exocentric cameras, whereas in MR settings, motion inference relies solely on the egocentric camera integrated into the HMD. In such cases, the most readily accessible information is limited to the user's head and hand trajectories~\cite{dittadi2021full}. Although approaches exist to reconstruct full-body motion from such sparse inputs~\cite{jiang2022avatarposer, ponton2023sparseposer, winkler2022questsim}, the estimated results are often noisy and less reliable. Noting these limitations, many prior studies have focused on interpreting user motion using only head and hand data for system efficiency and robustness~\cite{lee2023questenvsim, aliakbarian2022flag}. Our method aligns with this current, leveraging only the sparse motion cues from head and hand trajectories. This approach enhances system stability by not relying on unreliable full-body data, while also offering the significant advantage of adaptability. This allows for easy application across various MR scenarios and devices without the need for separate full-body tracking equipment.

An additional challenge in MR systems concerns the use of raw spatial data, which offers efficiency in both time and computational resources. Intention recognition has been frequently performed in an object-wise manner in previous studies, determining whether or not the user has an intention toward a specific object within the scene. Such object-wise methods~\cite{kratzer2020anticipating, clarence2021unscripted, huang2025human} require assessing each object in the user’s view to infer which object the user is interested in, necessitating the preemptive segmentation and tracking of objects within the scene. Although spatial recognition has been extensively studied with many efficient approaches available, applying inference across the entire scene inevitably introduces latency and overheads due to the limited computational resources of the HMD. Moreover, few studies~\cite{choi2022preemptive, leon2022intuitive, wei2017inferring, shi2021understand} that introduced spatial approaches have mainly been conducted in simple or single indoor environments. Therefore, we adopt a spatial approach that directly identifies intention within raw spatial data without additional processes and conduct experiments on both simple and challenging configuration, highlighting its robustness under real-world conditions.

Based on these considerations, we propose Int3DNet, a method that combines sparse motion with geometric scene representations to predict the intention area over a 3D scene. The sparse motion cues are fused with head orientation, which we use as a proxy for human attention instead of gaze, inspired by \cite{hu2024hoimotion}. While gaze is a strong cue for inferring intention~\cite{yin2024robust, pettersson2024intended, chen2022gaze}, it is not always available on HMDs. In contrast, head orientation can be readily obtained from HMD sensors and even be more informative than gaze cues for predicting future intentions~\cite{doshi2009roles}. Based on these findings, we adopt head orientation as a representation of human attention. The sparse motion cues and head orientation are transformed using a discrete cosine transform (DCT) and encoded using a graph-based spatio-temporal encoder. In parallel, the scene geometry, represented as a point cloud, is encoded using PointNet++~\cite{qi2017pointnet++}, which robustly encodes cluttered environments while maintaining invariance to point order. The sparse motion and point cloud encodings are fused via a cross attention to produce a 3D intention area within the scene. Given that neural attention mechanisms mimic aspects of human attention~\cite{lai2020understanding, das2017human}, we assume that the attention weights assigned to the scene points reflect human attention. Building on this assumption, we spatially model the 3D intention using cross attention between scene and motion features.

We evaluated the proposed method across different future time horizons of up to 1.5 seconds, using the MoGaze~\cite{kratzer2020mogaze} dataset for simple scene configurations and the CIRCLE~\cite{araujo2023circle} dataset for complex scene configurations. Gaze-based approaches, which areeffective in simple scenes like tabletop, become inapplicable when obstacles exist between user and target. Therefore, we conducted experiments not only on the MoGaze dataset but also on the challenging CIRCLE dataset to validate our method's ability across various environments. Our method was evaluated in comparison with prior approaches utilizing human attention, motion forecasting, and scene affordance alone. The evaluation metrics assessed how accurately the predicted intention area aligned with the ground-truth intention area. The results demonstrate that our method outperforms the baselines, achieving higher accuracy and overlap scores across both datasets and time horizons. This robustness highlights its clear benefit for MR systems, enabling the proactive prediction of user intention in real-world environments.
Furthermore, we demonstrate the usability and generalizability of our approach by implementing an intention-based Visual Question Answering (VQA) task on the Aria Digital Twins (ADT) dataset~\cite{pan2023aria}, which leverages a foundation vision-language model (VLM) to efficiently reason about the user’s future area of interest.

This study makes three major contributions. 
First, we propose a novel 3D intention prediction network for MR that integrates scene context and sparse motion cues, including head-hand trajectories and head orientations. Unlike previous studies, the network is designed for MR that accounts for its practical constraints.
Second, we validate the robustness of our method across diverse scene configurations, going beyond prior work limited to simple environments. 
Third, we demonstrate an intention-based VQA pipeline as a potential MR application leveraging the proposed network. This highlights the practical utility of our intention prediction method in real-world MR scenarios and its generalization capability.

\section{Related Work}

\subsection{Intention in Human-Computer Interaction (HCI)}
Human intention is a core concept in HCI, as interactive systems need to anticipate user goals to enable a seamless experience. Many researchers have explored diverse approaches to identify human interaction intent, ranging from physiological sensing~\cite{yang2023learning,zhao2020research} to behavioral modeling~\cite{bednarik2012you}. The importance of intention recognition extends to embodied interaction, leading to extensive research on human intention in HRI. Specifically, in human-robot collaborative manipulation tasks, intention prediction allows robots to proactively adapt their actions, ensuring both safety and responsiveness within shared workspace~\cite{mainprice2013human, choi2022preemptive, laplaza2024enhancing}. 
Furthermore, \cite{huang2025human} proposed a method for tele-grasping, where a remote robot manipulates objects. Their method uses human intention prediction to detect a target object and perform natural grasp planning for the object. 

Despite the similarities between MR and HRI, such as the need for systems to react adaptively and immediately to human interactions, the study of human intention in MR remains underexplored compared to that in HRI. While some studies~\cite{clarence2021unscripted, david2021towards, gamage2021so, chen2022gaze} have been conducted, most are limited to VR environments and have not considered real-world environment. 
To address this, we propose a human intention prediction method for MR that accounts for real-world scenes. 

\subsection{Intention Inference}
\begin{figure*}[tb]
  \centering 
  \includegraphics[width=\textwidth]{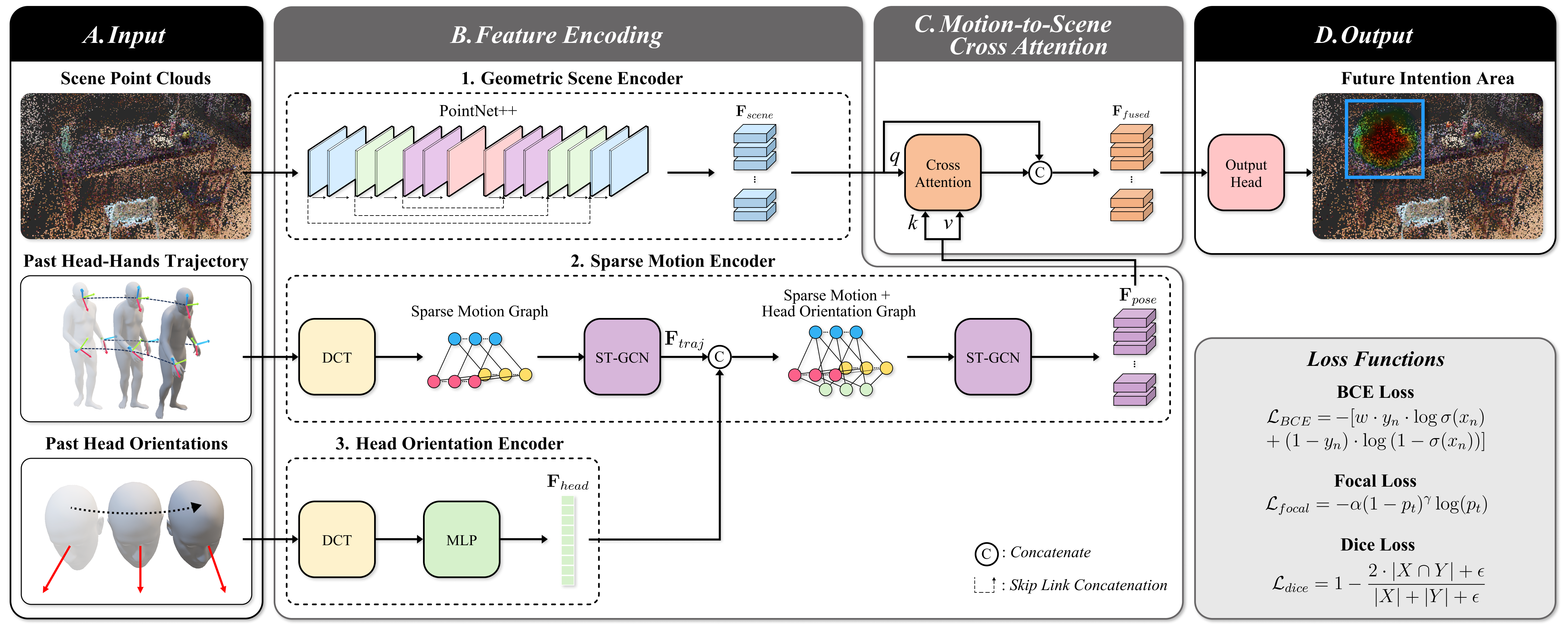}
  \caption{\textbf{Architecture of Int3DNet.} Scene point clouds, past head–hands trajectories, and head orientations are encoded via PointNet++, ST-GCN, and DCT–MLP modules. Motion and orientation features are fused with scene features through a cross attention, and the output head predicts the future intention area. The network is trained with a combination of weighted BCE, focal, and dice losses.}
  \vspace{-0.2cm}
  \label{fig:fig2}
\end{figure*}

Visual attention is closely related to human intention~\cite{land1999roles}. Visual attention is a cognitive process that allows a human to selectively focus on specific visual information while ignoring other distractions~\cite{goldstein2002sensation}.
It shifts according to human intention, and conversely, human intention can be interpreted using visual attention. 
Since visual attention is conveyed through eye movements~\cite{frischen2007gaze}, eye gaze is widely employed for intention inference~\cite{belardinelli2024gaze, bednarik2012you, david2021towards, yin2024robust}. 
Specifically, Bednarik et al.~\cite{bednarik2012you} proposed a framework for task-independent prediction of interaction intents for gaze-based interfaces. Yin et al.~\cite{yin2024robust} developed a gaze region model with density-based clustering for 3D object selection tasks.

Along with gaze, human motion is also a crucial cue that conveys human intention. Human intention can be interpreted by predicting future human motion from past human motion~\cite{luo2019human, liu2021motion, laplaza2024enhancing}. Otherwise, apart from approaches that rely on predicting future motion, Yu et al.~\cite{yu2015human} proposed an approach that classifies motion labels from observed motion, and then predicts intention based on the classified labels. Kratzer et al.\cite{kratzer2020anticipating} inferred intention by estimating affordances from past motion. Using human motion as a cue allows the system to address implicit intentions not observable through gaze, while also capturing spatial intention by incorporating user movements in 3D environments. While existing approaches use full pose~\cite{kratzer2020anticipating, shi2021understand} or upper-limb pose~\cite{gamage2021so, kedia2024interact, laplaza2022context}, these are limited in MR environments. Accordingly, we propose an intention prediction method that utilizes only head-hand motion input. Additionally, several gaze and motion fused studies~\cite{yan2023gazemodiff, chen2022gaze, yeamkuan20213d, he2018real} indicate that the collaborative use of gaze and motion is more representative than single use. Following this insight, we utilized both gaze and sparse motion inputs. However, as gaze tracking is not always available on HMDs and head orientation provides a more reliable cue for predicting future intentions~\cite{doshi2009roles, doshi2012head}, we adopted head orientation instead of eye gaze. 

\subsection{Intention Representation}
The representation of human intention varies depending on the domain and task. Various approaches have been proposed to represent human intention in prior studies. Among them, the concept of saliency has been widely employed in computer vision and HCI research. A saliency map typically denotes the distribution of a user's visual attention over a 2D image plane, often represented in the form of a heatmap~\cite{hayes2021deep, chang2023human}. Gaze-based intention is also widely represented as a 2D representation; Koochaki et al.~\cite{koochaki2019eye} identify target objects seen in a displayed 2D image using both spatial and temporal information from the eye gaze. However, the 2D representation is sensitive to occlusion and cannot represent the intentions of target in position not visible from the camera's perspective. 

To address these limitations, human intention has been represented in 3D space~\cite{leon2022intuitive, choi2022preemptive, hu2022we, shi2021understand, wei2017inferring}. Specifically, Leon et al.~\cite{leon2022intuitive} enabled collaborative tasks with a robot arm by leveraging upper limb 3D motion by predicting intention area on tabletop. Some studies~\cite{wei2017inferring, shi2021understand} explored intention prediction in real indoor scenes, proposing methods that utilize human skeleton poses to estimate intention areas within 3D environments. However, despite the existing studies on 3D intention, consideration for diverse and complex scenes remain underexplored, which is a critical factor for deployment in real-world MR. Therefore, we conducted a comprehensive evaluation to verify whether the proposed method remains robust across various scene configurations.

\section{Method}
\subsection{Problem Definition}
We define intention prediction as estimating a spatial heatmap over a scene point cloud, where each point has a score indicating the likelihood of future user interaction. Formally, given a scene $\mathbf{S} \in \mathbb{R}^{N \times 3}$, where $N$ denotes the number of spatial points, the goal is to predict as intention score $i \in \mathbb{R}^{N}$ such that higher values indicate regions of expected interaction (e.g. area where the interaction target is placed or will be placed). Capturing object segmentation can be challenging in an MR environment due to dynamic real-world scenes. However, through real-time depth sensing and 3D reconstruction, a fused scene point cloud can be acquired from HMDs. Accordingly, we define our problem as 3D intention area prediction, aiming to predict intention regions directly from the scene point cloud $\mathbf{S}$ without explicit object-level perception.

To infer intention, we used sparse body motion and head orientation as primary cues. Specifically, we used the 3D positions of the head and both hands $\mathbf{j}_t \in \mathbb{R}^{3 \times 3}$ and their linear velocities $\mathbf{v}_t \in \mathbb{R}^{3 \times 3}$ at timestamp $t$. We only used the head and both hands because they are readily available information from recent commercial HMD. In addition, we introduce the use of the head orientation vector $\mathbf{h}_t \in \mathbb{R}^{3}$ as a proxy for gaze, where $\mathbf{h}_t$ is a unit vector indicating the forward direction of the head. 
Specifically, given a temporal sequence of length $T$ past hand and head poses and velocities $\mathbf{M}_{1:T} = [\mathbf{j}_1, \ldots, \mathbf{j}_T, \mathbf{v}_1, \ldots, \mathbf{v}_T]$, head orientation vectors $\mathbf{H}_{1:T} = [\mathbf{h}_1, \ldots, \mathbf{h}_T]$, and a scene point cloud $\mathbf{S}$, the network outputs a 3D intention heatmap $i \in \mathbb{R}^{N}$ where an interaction will probabilistically occur.

\subsection{Feature Encoding}
The input point cloud $\mathbf{S} \in \mathbb{R}^{N \times 3}$ is first passed through a PointNet++~\cite{qi2017pointnet++} encoder-decoder architecture that hierarchically abstracts and reconstructs the spatial structure~(\cref{fig:fig2}~(B-1)). The encoder processes $\mathbf{S}$ into multi-scale latent features via sampling and grouping operations. The decoder then upsamples these features back to per-point embeddings, resulting in a scene feature matrix $\mathbf{F}_{scene} \in \mathbb{R}^{N \times D}$, where $D$ is the feature dimension.

To represent user motion, we extract 3D positions $\mathbf{j}_t$ and velocities $\mathbf{v}_t$ of a sparse set of joints (head and both hands) across the past $T$ frames. These are concatenated into a spatio-temporal representation $\mathbf{M}_{1:T} \in \mathbb{R}^{T \times 3 \times 6}$, where each element contains both the 3D position and its corresponding velocity. As shown in \cref{fig:fig2}~(B-2), we then apply a Discrete Cosine Transform (DCT) to reduce temporal redundancy and highlight dominant motion frequencies~\cite{hu2024hoimotion, ma2022progressively}. The DCT of a sequence $\mathbf{z} \in \mathbb{R}^{T \times d}$ is defined as:
\begin{equation}\hat{\mathbf{z}}[k] = \sum_{t=0}^{T-1} \mathbf{z}[t] \cdot \cos\left( \frac{\pi}{T} \left(t + \frac{1}{2} \right) k \right), \quad k = 0, \ldots, T-1\end{equation}

\noindent which is applied along the temporal axis. The transformed motion sequence  $\hat{\mathbf{M}}$  is encoded via a spatio-temporal graph convolutional network (GCN)  encoder $f_{\text{pose}}^{enc}$ to yield latent motion features $\mathbf{F}_{traj} \in \mathbb{R}^{T  \times 3 \times D}$.
In parallel, the head orientation sequence $\mathbf{H}_{1:T}$ is also transformed with DCT and passed through an MLP to obtain head orientation motion embeddings $\mathbf{F}_{head} \in \mathbb{R}^{T \times 1 \times D}$~(\cref{fig:fig2}~(B-3)). These features are concatenated along the joint axis, forming $\mathbf{F}_{motion} = [\mathbf{F}_{traj}, \mathbf{F}_{head}] \in \mathbb{R}^{T  \times 4 \times D}$. Lastly, the concatenated motion feature is decoded using another spatio-temporal GCN $f_{\text{pose}}^{dec}$ to obtain the final motion representation $\mathbf{F}_{pose} \in \mathbb{R}^{T \times D}$ after joint-axis pooling~(\cref{fig:fig2}~(B-2)).

\subsection{Motion-to-Scene Cross-Attention}
To condition the scene representation on user motion, we applied a linear cross attention mechanism~\cite{katharopoulos2020transformers}~(\cref{fig:fig2}~(C)). Let the scene features $\mathbf{F}_{scene} \in \mathbb{R}^{N \times D}$ serve as the query $\mathbf{Q}$, while the temporally encoded motion features $\mathbf{F}_{pose} \in \mathbb{R}^{T \times D}$ serve as both key $\mathbf{K}$ and value $\mathbf{V}$. The attention output is computed as:
\begin{equation}
\text{Attention}(\mathbf{Q}, \mathbf{K}, \mathbf{V})_i = 
\frac{\phi(Q_i)\left(\sum_{j=1}^T \phi(K_j)^\top V_j \right)}
     {\phi(Q_i)\left(\sum_{j=1}^T \phi(K_j)^\top \right)},
\end{equation}
where $\phi(\cdot)$ represents the kernel function. This results in $\mathbf{A}_{pose} \in \mathbb{R}^{N \times D}$, which is concatenated with the original scene features to produce the fused representation $\mathbf{F}_{fused} = [\mathbf{F}_{scene}, \mathbf{A}_{pose}] \in \mathbb{R}^{N \times 2D}$. The fused representation is passed through an MLP output head that projects the $2D$-dimensional fused features to intention scores for each point $i \in \mathbb{R}^{N}$~(\cref{fig:fig2}~(D)). The resulting $i$ represents the predicted intention heatmap, where each value corresponds to the probability of a user interaction at a specific point in the scene.

\subsection{Loss Function}
To effectively supervise the prediction of 3D spatial intention over point clouds, we adopted a composite loss function that jointly addressed the challenges of class imbalance, sparse positive regions, and spatial coherence. The total loss \(\mathcal{L}_{total}\) combines a weighted binary cross-entropy loss \(\mathcal{L}_{BCE}\), focal loss \(\mathcal{L}_{focal}\) and dice loss \(\mathcal{L}_{dice}\), each contributing complementary strengths to the training objective:
\begin{equation}\mathcal{L}_{total} = \mathcal{L}_{BCE} + \mathcal{L}_{focal} + \mathcal{L}_{dice}. \end{equation}

Given the extreme imbalance between the interest and background areas within the point cloud, we apply a class-weighted binary cross-entropy loss \(\mathcal{L}_{BCE}\), where the positive class is up-weighted proportionally to the ratio between the negative and positive samples in the dataset:
\begin{equation}\mathcal{L}_{BCE} = -[w \cdot y_{n} \cdot \log{\sigma (x_{n}) + (1-y_{n})\cdot\log{(1-\sigma(x_n))} }],\end{equation}
where \(N\) denotes the total number of points, \(x_n\) is the raw model output for the n-th point, and \(y_n \in \{0,1\}\) is the corresponding ground-truth binary label. \(\sigma(\cdot)\) represents the sigmoid activation function, and w is the class weight applied to the positive class to compensate for the imbalance between intention and non-intention points. Inspired by \cite{nguyen2023open}, the class weight w is defined as
\begin{equation}w = \frac{N_{neg}}{N_{pos}},\end{equation}
where \(N_{neg}\) and \(N_{pos}\) denote the numbers of negative and positive points, respectively. 

To further enhance learning from hard examples, we incorporate a focal loss~\cite{lin2017focal} $\mathcal{L}_{focal}$, which modulates the contribution of each point based on the prediction confidence:
\begin{equation}\mathcal{L}_{focal} = - \alpha (1 - p_t) ^ \gamma \log(p_t),\end{equation}
where $p_t$ is the predicted probability of the true class, $\alpha$ is a balancing factor, and $\gamma$ is a focusing parameter. 

In addition, to promote region-level consistency and encourage overlap between the predicted and ground-truth intention areas, we also applied a dice loss~\cite{milletari2016v} $\mathcal{L}_{dice}$ term:
\begin{equation}\mathcal{L}_{dice} = 1 - \frac{2 \cdot |X \cap Y| + \epsilon}{|X| + |Y| + \epsilon},\end{equation}
where $X$ and $Y$ are the predicted and ground-truth binary masks, respectively, and $\epsilon$ is $10^{-6}$. 

\section{Experiments}
\subsection{Dataset}
\begin{figure}[tb]
  \centering 
  \includegraphics[width=0.9\columnwidth]{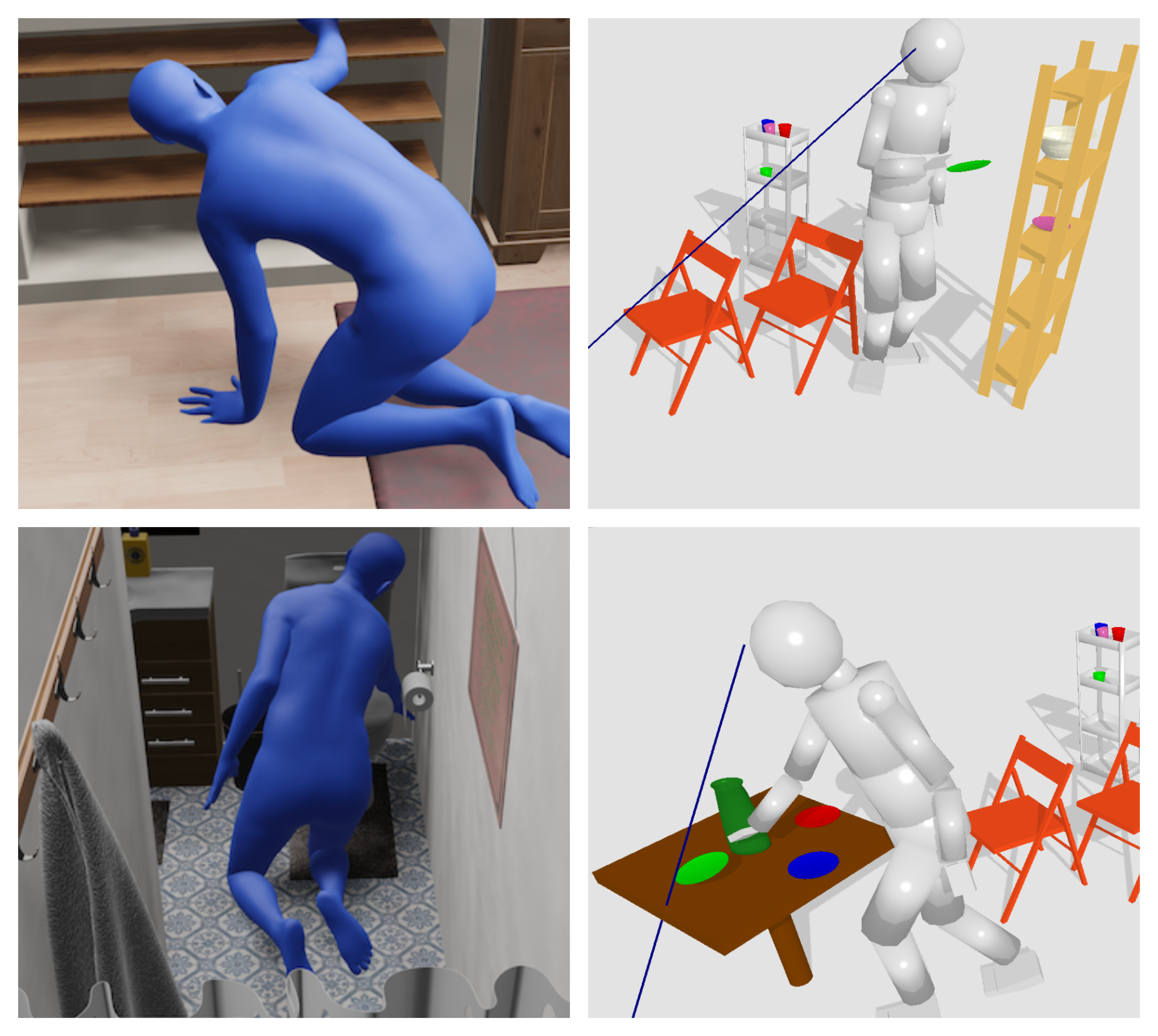}
  \caption{\textbf{Examples from two datasets.} The CIRCLE dataset (left column) contains complex scene configurations, while the MoGaze dataset (right column) contains simpler scenes. 
  Reprinted from \cite{araujo2023circle} and \cite{kratzer2020mogaze}. }
  \vspace{-0.4cm}
  \label{fig:fig3}
\end{figure}

We conducted comparative experiments on the MoGaze and CIRCLE datasets for the 3D intention prediction task. The MoGaze dataset~\cite{kratzer2020mogaze} is a collection of human motion data for pick-and-place tasks in an indoor environment. It includes full-body motion capture data, 3D scene geometry, and gaze. It comprises approximately 180 minutes of motion data collected from seven participants performing manipulation actions. The CIRCLE dataset~\cite{araujo2023circle} is a large-scale collection of 3D human motion captured in diverse virtual reality environments. It contains 10 hours of full-body reaching motion from five participants interacting with cluttered indoor environments across nine scenes. 

As shown in \cref{fig:fig3}, The MoGaze dataset contains interactions where goal objects are easily visible, while CIRCLE includes cluttered scenes with partially or fully occluded goal positions. In real-world MR environments, both open and cluttered scene naturally exists. Therefore, the intention prediction network need to be robust across such diverse spatial contexts. However, prior approaches~\cite{choi2022preemptive, leon2022intuitive} were mainly evaluated in simple scenes like tabletop environments, where gaze cues alone could identify the user's intended area. In cluttered scenes like the CIRCLE dataset, gaze cues do not always correspond to the actual intention area, highlighting the need for 3D spatial intention prediction. To address this, we conducted experiments on both the MoGaze and CIRCLE datasets, which differ in scene complexity, showing our proposed network performs robustly across varying configurations. Additionally, we adopted a scene-based split with mutually exclusive training and test scenes to evaluate generalization to unseen environments. This setup provides a more challenging evaluation setting and meaningful benchmark for future intention prediction research.

The MoGaze and CIRCLE datasets both consist of multiple action segments within a dataset but differ in how segments are defined. In the MoGaze dataset, multiple action segments involve within a single session. Thus, we extracted $T$-second windows, one prior to the moment the user grasped the object and another prior to the moment the object was placed, from a single session. In contrast, the CIRCLE dataset provides segmented reaching motions, each recorded from the initiation of a reach to the moment of reaching the goal position. For our experiments, we extracted $T$-second windows prior to reaching for each segment. For both datasets, we aligned human pose sequences and scene geometry based on body orientation at $t=0$. Scene point clouds were sampled from the surfaces of both static and dynamic object meshes using a surface sampling. To simulate realistic 3D point clouds as acquired by HMD depth sensors, we sampled a fixed number of scene points using a distance-weighted strategy that prioritizes points closer to the user. For ground-truth 3D intention area annotation, we generated a 3D Gaussian heatmap based on distance from goal object or placement position. The ground truth intention area was then defined as a subset of points whose heatmap values exceeded a threshold.

\subsection{Baseline}
To evaluate the performance of Int3DNet, we conducted comparative analysis with five baselines. 
\begin{itemize}[itemsep=1pt, topsep=2pt, parsep=2pt, leftmargin=1em]
  \item \textit{Head Orientation} Model: 
This model predicts intention by projecting a ray from the head-facing vector as a proxy for gaze. 
According to \cite{kratzer2020mogaze}, gaze distance methods outperform skeleton information or neural networks.

  \item \textit{Head+Scene} Model: This model combines head orientation with scene geometry to predict the intention area, integrating geometric cues for more context-aware prediction rather than simply casting a ray from head direction.

  \item \textit{Motion Forecast} Model~\cite{hu2024hoimotion}: 
This model employs a motion forecasting network to predict the user's future motion trajectory from past motion frames. The intention area is defined as the region surrounding the wrist of the interacting hand in the predicted future pose. 

  \item \textit{Only Scene} Model: This model predicts user intention using only the geometric information of the scene. It evaluates whether intention areas can be inferred from scene affordances alone.
\end{itemize}
We further conducted ablation studies on the attention mechanism:
\begin{itemize}[itemsep=1pt, topsep=2pt, parsep=2pt, leftmargin=1em]
  \item \textit{MLP-based} Model: This model replaces the attention module with a simple multi-layer perceptron, fusing pose, gaze, and scene features without cross attention mechanism.
  \item \textit{Motion Query} Model: This model uses body motion as queries and scene features as keys/values to test the effect of attention directionality.
\end{itemize}
Additionally, we performed ablation studies on the loss functions by comparing the effects of BCE, Focal, and Dice losses.

\subsection{Metrics}
To evaluate the performance of our proposed method for 3D spatial intention prediction, we used four metrics: Similarity (SIM), Area Under Curve (AUC), mean Intersection over Union (mIOU), and Dice score. Previous studies focused on object-centric intention prediction, using accuracy of selecting the user's intended object from choices as the primary metric~\cite{kratzer2020anticipating, clarence2021unscripted, huang2025human}. However, these metrics are insufficient to evaluate our approach, as our method predicts intention as a heatmap within the 3D spatial scene. We employed metrics common in 3D spatial localization tasks~\cite{shao2025great, wang2025grace} for evaluation. SIM measures distributional alignment between predicted and ground-truth heatmaps via normalized histogram intersection. AUC assesses the model's ability to rank intention versus non-intention regions without thresholding. mIoU quantifies spatial overlap by averaging binary IoU scores over multiple thresholds. While Dice loss is used during training to promote region-level consistency, the Dice score serves as a quantitative measure of predicted and ground-truth heatmap alignment.

\section{Results}
\subsection{Quantitative and Qualitative Results}

\begin{table*}[tb]
  \caption{\textbf{Quantitative results on the CIRCLE dataset.} The performance was reported across different time horizons (500–1500 ms before interaction) using four metrics: SIM, AUC, mIoU, and Dice. Our method consistently outperforms all baselines across all metrics and time horizons.}
  \label{tab:tab1}
\centering
\resizebox{0.9\textwidth}{!}{
\begin{tabular}{ccccccccccccccccc}
\toprule
Metrics& Method & 500 ms & 600 ms & 700 ms & 800 ms & 900 ms & 1000 ms & 1100 ms & 1200 ms & 1300 ms & 1400 ms & 1500 ms & Average\\ \hline
\multirow{5}{*}{\textit{SIM} $\uparrow$}
  &\textit{Head Orientation} & 0.1791 & 0.1781 & 0.2017 & 0.1854 & 0.2024 & 0.2042 & 0.2022 & 0.1989 & 0.1930 & 0.1966 & 0.2040 & 0.1951 \\
  &\textit{Head+Scene} & 0.2173 & 0.2175 & 0.2176 & 0.2178 & 0.2180 & 0.2183 & 0.2184 & 0.2185 & 0.2183 & 0.2180 & 0.2175 & 0.2179 \\
  &\textit{Motion Forecast} & 0.3793 & 0.3771 & 0.3715 & 0.3643 & 0.3529 & 0.3432 & 0.3353 & 0.3167 & 0.3004 & 0.2872 & 0.2752 & 0.3367 \\
  &\textit{Only Scene} & 0.1770 & 0.1770 & 0.1770 & 0.1770 & 0.1770 & 0.1770 & 0.1770 & 0.1770 & 0.1770 & 0.1770 & 0.1770 & 0.1770 \\
  &Ours & \textbf{0.4107} & \textbf{0.4084} & \textbf{0.4072} & \textbf{0.4041} & \textbf{0.3989} & \textbf{0.3937} & \textbf{0.3886} & \textbf{0.3824} & \textbf{0.3770} & \textbf{0.3726} & \textbf{0.3680} & \textbf{0.3920} \\
  \midrule

  \multirow{5}{*}{\textit{AUC} $\uparrow$}
  &\textit{Head Orientation} & 83.88 &	84.50 &	86.50 &	85.81 &	87.47 &	87.36 &	87.61 &	86.80 &	86.96 &	86.84 &	87.66 &	86.49  \\
  &\textit{Head+Scene} & 93.35 &	93.35 &	93.35 &	93.35 &	93.36 &	93.36 &	93.36 &	93.36 &	93.35 &	93.33 &	93.31 &	93.35 \\
  &\textit{Motion Forecast} & 88.92 &	88.93 &	88.89 &	88.84 &	88.76 &	88.74 &	88.67 &	88.40 &	88.20 &	88.07 &	87.85 &	88.57 \\
  &\textit{Only Scene} & 92.22 &	92.22 &	92.22 &	92.22 &	92.22 &	92.22 &	92.22 &	92.22 &	92.22 &	92.22 &	92.22 &	92.22  \\
  &Ours & \textbf{97.69} & \textbf{97.67} & \textbf{97.65} & \textbf{97.63} & \textbf{97.58} & \textbf{97.50} & \textbf{97.43} & \textbf{97.35} & \textbf{97.27} & \textbf{97.21} & \textbf{97.12} & \textbf{97.46} \\
  \midrule

  \multirow{5}{*}{\textit{mIoU} $\uparrow$}
  &\textit{Head Orientation} & 6.35 & 6.25 & 7.21 & 6.58 & 7.43 & 7.36 & 7.16 & 7.03 & 7.22 & 6.81 & 7.20 & 6.96 \\
  &\textit{Head+Scene} & 16.07 & 16.08 & 16.09 & 16.10 & 16.12 & 16.13 & 16.14 & 16.13 & 16.09 & 16.04 & 15.94 & 16.08 \\
  &\textit{Motion Forecast} & 15.90 & 15.69 & 15.49 & 15.22 & 14.56 & 13.99 & 13.50 & 12.60 & 11.82 & 11.11 & 10.61 & 13.68 \\
  &\textit{Only Scene} & 13.57 & 13.57 & 13.57 & 13.57 & 13.57 & 13.57 & 13.57 & 13.57 & 13.57 & 13.57 & 13.57 & 13.57\\
  &Ours & \textbf{27.85} & \textbf{27.72} & \textbf{27.60} & \textbf{27.36} & \textbf{27.00} & \textbf{26.57} & \textbf{26.15} & \textbf{25.75} & \textbf{25.42} & \textbf{25.15} & \textbf{24.90} & \textbf{26.50} \\
  \midrule

  \multirow{5}{*}{\textit{Dice} $\uparrow$}
  &\textit{Head Orientation} & 0.0941 & 0.0930 & 0.1089 & 0.0944 & 0.1118 & 0.1060 & 0.1056 & 0.1009 & 0.1120 & 0.1055 & 0.1146 & 0.1043 \\
  &\textit{Head+Scene} & 0.3102 & 0.3105 & 0.3105 & 0.3106 & 0.3109 & 0.3115 & 0.3119 & 0.3120 & 0.3118 & 0.3108 & 0.3085 & 0.3108 \\
  &\textit{Motion Forecast} & 0.2312 & 0.2300 & 0.2267 & 0.2256 & 0.2164 & 0.2108 & 0.2051 & 0.1878 & 0.1776 & 0.1662 & 0.1612 & 0.2035 \\
  &\textit{Only Scene} & 0.2668 & 0.2668 & 0.2668 & 0.2668 & 0.2668 & 0.2668 & 0.2668 & 0.2668 & 0.2668 & 0.2668 & 0.2668 & 0.2668 \\
  &Ours & \textbf{0.4435} & \textbf{0.4422} & \textbf{0.4415} & \textbf{0.4380} & \textbf{0.4349} & \textbf{0.4303} & \textbf{0.4235} & \textbf{0.4179} & \textbf{0.4144} & \textbf{0.4112} & \textbf{0.4067} & \textbf{0.4276} \\
  \bottomrule        
\end{tabular}} 
\end{table*}

\begin{table*}[tb]
  \caption{\textbf{Quantitative results on the MoGaze dataset.} Performance is reported across different time horizons (500–1500 ms before interaction) using four metrics: SIM, AUC, mIoU, and Dice. Our method outperforms all baselines across all metrics up to 1200 ms, while beyond this horizon the performance advantage diminishes due to the characteristics of the dataset.}\label{tab:tab2}
\centering
\resizebox{0.9\textwidth}{!}{
\begin{tabular}{ccccccccccccccccc}
\toprule
Metrics& Method & 500 ms & 600 ms & 700 ms & 800 ms & 900 ms & 1000 ms & 1100 ms & 1200 ms & 1300 ms & 1400 ms & 1500 ms & Average\\ \hline
\multirow{5}{*}{\textit{SIM} $\uparrow$}
  &\textit{Head Orientation} & 0.1266 & 0.1271 & 0.1246 & 0.1243 & 0.1131 & 0.1060 & 0.0997 & 0.0850 & 0.0691 & 0.0570 & 0.0506 & 0.0985 \\
  &\textit{Head+Scene} & 0.3051 & 0.3048 & 0.3047 & 0.3047 & 0.3047 & 0.3047 & 0.3046 & 0.3044 & 0.3044 & 0.3046 & 0.3047 & 0.3047 \\
  &\textit{Motion Forecast} &0.1529 & 0.1598 & 0.1863 & 0.2346 & 0.2699 & 0.2499 & 0.2132 & 0.1970 & 0.1813 & 0.1698 & 0.1601 & 0.1977 \\
  &\textit{Only Scene} & 0.3053 & 0.3053 & 0.3053 & 0.3053 & 0.3053 & 0.3053 & 0.3053 & 0.3053 & 0.3053 & 0.3053 & \textbf{0.3053} & 0.3053 \\
  &Ours & \textbf{0.4637} & \textbf{0.4327} & \textbf{0.3947} & \textbf{0.3662} & \textbf{0.3471} & \textbf{0.3340} & \textbf{0.3205} & \textbf{0.3145} & \textbf{0.3105} & \textbf{0.3077} & 0.3036 & \textbf{0.3541} \\
  \midrule

  \multirow{5}{*}{\textit{AUC} $\uparrow$}
  &\textit{Head Orientation} & 65.54 & 65.80 & 65.71 & 65.98 & 65.01 & 64.25 & 63.38 & 61.79 & 60.20 & 58.80 & 56.82 & 63.03 \\
  &\textit{Head+Scene} & 83.67 & 83.63 & 83.60 & 83.57 & 83.56 & 83.52 & 83.43 & 83.39 & 83.41 & \textbf{83.43} & \textbf{83.45} & 83.51 \\
  &\textit{Motion Forecast} & 80.69 & 80.24 & 81.43 & 84.10 & 85.69 & 84.72 & 83.22 & 82.44 & 81.74 & 80.80 & 79.57 & 82.24 \\
  &\textit{Only Scene} & 83.27 & 83.27 & 83.27 & 83.27 & 83.27 & 83.27 & 83.27 & 83.27 & 83.27 & 83.27 & 83.27 & 83.27 \\
  &Ours & \textbf{91.69} & \textbf{91.00} & \textbf{89.46} & \textbf{87.95} & \textbf{86.71} & \textbf{85.34} & \textbf{84.63} & \textbf{84.15} & \textbf{83.56} & 83.04 & 82.59 & \textbf{86.38} \\
  \midrule

  \multirow{5}{*}{\textit{mIoU} $\uparrow$}
  &\textit{Head Orientation} & 5.12 & 5.26 & 5.24 & 5.01 & 4.79 & 4.41 & 4.30 & 3.69 & 3.47 & 2.84 & 2.60 & 4.25 \\
  &\textit{Head+Scene} & 27.36 & 27.33 & 27.31 & 27.31 & 27.33 & 27.33 & 27.33 & 27.32 & 27.33 & \textbf{27.35} & \textbf{27.38} & 27.33 \\
  &\textit{Motion Forecast} & 4.13 & 4.70 & 5.77 & 7.79 & 9.09 & 7.97 & 6.69 & 5.57 & 5.09 & 4.56 & 4.28 & 5.97 \\
  &\textit{Only Scene} & 27.35 & 27.35 & 27.35 & 27.35 & 27.35 & 27.35 & 27.35 & 27.35 & \textbf{27.35} & 27.35 & 27.35 & 27.35 \\
  &Ours & \textbf{39.05} & \textbf{36.93} & \textbf{34.02} & \textbf{31.77} & \textbf{30.28} & \textbf{29.05} & \textbf{28.01} & \textbf{27.62} & 27.09 & 26.73 & 26.40 & \textbf{30.63} \\
  \midrule

  \multirow{5}{*}{\textit{Dice} $\uparrow$}
  &\textit{Head Orientation} & 0.0764 & 0.0805 & 0.0761 & 0.0712 & 0.0664 & 0.0612 & 0.0550 & 0.0410 & 0.0318 & 0.0200 & 0.0227 & 0.0548 \\
  &\textit{Head+Scene} & 0.4270 & 0.4268 & 0.4267 & 0.4267 & 0.4266 & 0.4264 & 0.4261 & 0.4260 & \textbf{0.4261} & \textbf{0.4263} & \textbf{0.4265} & 0.4265 \\
  &\textit{Motion Forecast} & 0.0320 & 0.0385 & 0.0491 & 0.0766 & 0.1026 & 0.0983 & 0.0757 & 0.0648 & 0.0515 & 0.0436 & 0.0409 & 0.0612 \\
  &\textit{Only Scene} & 0.4231 & 0.4231 & 0.4231 & 0.4231 & 0.4231 & 0.4231 & 0.4231 & 0.4231 & 0.4231 & 0.4231 & 0.4231 & 0.4231 \\
  &Ours & \textbf{0.5980} & \textbf{0.5644} & \textbf{0.5269} & \textbf{0.4885} & \textbf{0.4647} & \textbf{0.4501} & \textbf{0.4331} & \textbf{0.4312} & 0.4228 & 0.4195 & 0.4144 & \textbf{0.4740} \\
  \bottomrule        
\end{tabular}} 
\end{table*}

\noindent\textbf{Results on CIRCLE}
\cref{tab:tab1} shows the comparison results of evaluation metrics on the CIRCLE dataset. The experimental results show the prediction performance across different time horizons using 15-frame inputs sampled at 500 ms, 600 ms, 700 ms, and up to 1500 ms prior to the interaction, along with the average performance. Across all evaluation metrics, our proposed Int3DNet consistently outperformed the baselines (\textit{Head Orientation}, \textit{Head+Scene}, \textit{Motion Forecast}, and \textit{Only Scene}). Although the proposed method showed a gradual decrease in prediction performance as the input sampling point was shifted further back in time from the interaction, it consistently achieved higher accuracy than the baselines. \textit{Only Scene} and \textit{Head+Scene} maintained relatively stable performance across different sampling times, whereas \textit{Motion Forecast} showed a strong sensitivity to changes in the sampling times, showing substantial performance degradation as the sampling point shifted. Meanwhile, \textit{Head Orientation} remained unstable across horizons.

\noindent\textbf{Results on MoGaze}
\cref{tab:tab2} shows the comparison results on the MoGaze dataset. As in the CIRCLE dataset, we evaluated prediction performance at different time horizons, using 15-frame inputs sampled at 500 ms to 1500 ms prior to the interaction, together with the average performance. Similar to CIRCLE, our method exhibited a performance degradation as the temporal distance from the interaction increased. While our proposed Int3DNet consistently outperformed all baselines across evaluation metrics up to 1200 ms, its performance beyond this horizon became comparable or even lower than the baselines. This performance degradation is likely related to the characteristics of the MoGaze dataset, which are discussed in more detail in \cref{sec:discussion_1}.

\begin{figure}[tb]
  \centering 
  \includegraphics[width=\columnwidth]{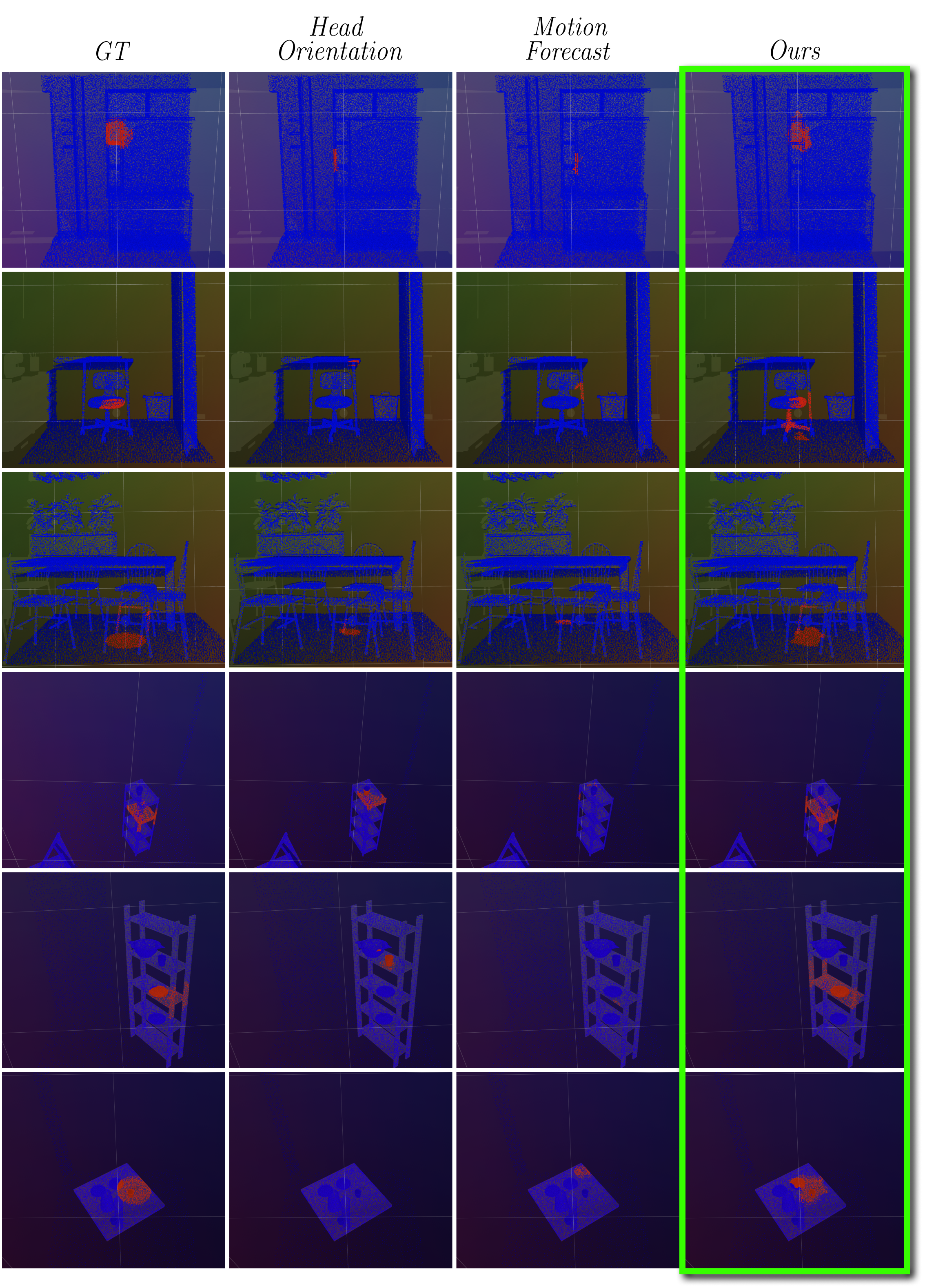}
  \caption{\textbf{Qualitative results of intention prediction.} The columns show, from left to right, the ground truth, \textit{Head Orientation} model, \textit{Motion Forecast} model, and our method. Rows 1–3 correspond to the CIRCLE dataset, and rows 4–6 to the MoGaze dataset.}
  \vspace{-0.4cm}
  \label{fig:fig4}
\end{figure}

\noindent\textbf{Qualitative Results}
\cref{fig:fig4} shows predicted 3D intention areas on the CIRCLE and MoGaze datasets, obtained 1500 ms and 1000 ms before the interaction, respectively. The red regions in the scene denote the 3D intention areas. Compared to the baselines, our method reliably localizes the intention area, whereas the baselines often include irrelevant regions or fail to identify the intention area at all. Additional results are provided in the Supplementary Video.

\subsection{Ablation Study}
We conducted ablation studies on our method to evaluate the impact of (1) the cross-attention mechanism between the scene and human sparse motion, (2) attention directionality, and (3) the loss function. 

\pagebreak

\noindent\textbf{Impact of cross attention mechanism} To evaluate the impact of the attention mechnism on model performance, we conducted an ablation study comparing a MLP model and the proposed method. \cref{tab:tab3}, shows the results in terms of SIM, AUC, mIOU, and Dice. Our method outperformed the MLP model across all metrics, indicating that explicitly modeling the pairwise relationships between scene geometry and human motion leads to more accurate predictions of spatial intention areas.

\noindent\textbf{Impact of attention directionality} In our method, the scene geometry features are used as queries, while the fused human motion and head orientation features serve as keys and values. To examine the impact of attention directionality, we also conducted an experiment in which queries were taken from the fused motion-head orientation and key/values from the scene geometry. The results in \cref{tab:tab3} show that using scene features as queries yields higher performance. When each scene point serves as a query, the attention weights are computed with respect to the fused motion cues for that specific spatial point. In contrast, when queries are derived from sequential motion and key/values from scene geometry, the attention mechanism attempts to capture the intention area from the entire scene, leading to high weights being assigned to distractor regions in cluttered environments. 

\begin{table}[t]
	\centering
        \caption{\textbf{Ablation results on the CIRCLE and MoGaze datasets.} Performance is reported as the average across all time horizons using four metrics: SIM, AUC, mIoU, and Dice.}\label{tab:tab3}
	\resizebox{\columnwidth}{!}{
	\begin{tabular}{cccccc}
		\toprule
		Dataset & Method & SIM $\uparrow$ & AUC $\uparrow$ & mIoU $\uparrow$ & Dice $\uparrow$\\ \hline
	\multirow{7}{*}{\textit{CIRCLE}}
        & Ours w/ MLP & 0.1991 & 93.07 & 14.03 & 0.2758\\
        & Ours w/ Motion Query & 0.1707 & 92.41 & 13.39 & 0.2659\\
        & Ours w/ BCE & 0.3398 & 97.27 & 25.84 & 0.4531 \\
        & Ours w/ Focal & 0.1295 & 96.58 & 9.74 & 0.0896\\
        & Ours w/ Dice & \textbf{0.4284} & 87.02 & \textbf{36.45} & \textbf{0.5279}\\
        & Ours w/ Dice+Focal & 0.4064 & 95.44 & 26.88 & 0.4305  \\
        & Ours & 0.3920 & \textbf{97.46} & 26.50 & 0.4276\\
        \midrule
        
	\multirow{7}{*}{\textit{MoGaze}} 
      & Ours w/ MLP & 0.3002 & 83.50 & 27.56 & 0.4262\\
      & Ours w/ Motion Query & 0.3097 & 83.31 & 27.77 & 0.4299\\
      & Ours w/ BCE & 0.3081 & 85.81 & 28.43  &0.4337 \\
      & Ours w/ Focal & 0.2711 & 83.26 & 15.52 & 0.1220 \\
      & Ours w/ Dice & 0.3737 & 77.41 & \textbf{30.87} & 0.4481 \\
      & Ours w/ Dice+Focal & \textbf{0.3810} & 83.90 & 26.42 & 0.4350 \\
      & Ours & 0.3541 & \textbf{86.38} & 30.63 & \textbf{0.4740}\\
        \bottomrule
	\end{tabular}}
  \vspace{-0.4cm}
\end{table}

\begin{figure*}[tb]
  \centering 
  \includegraphics[width=\textwidth]{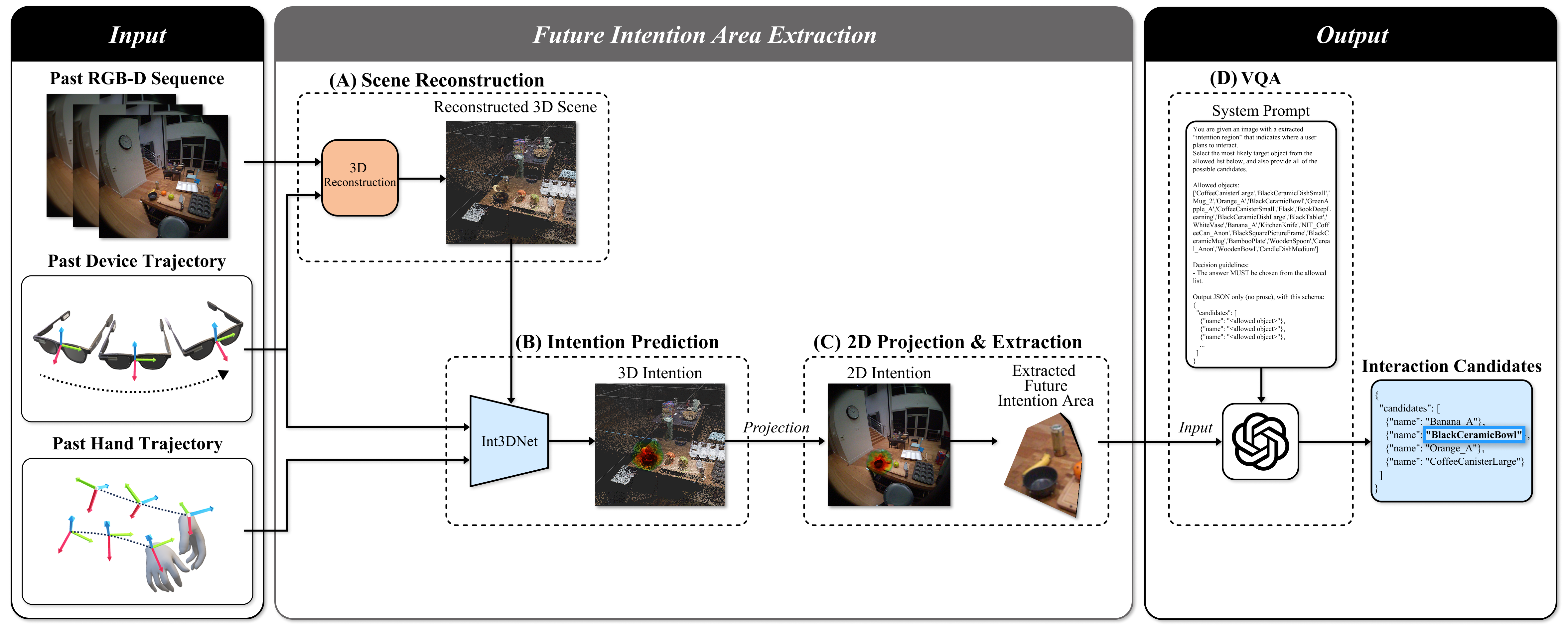}
  \caption{\textbf{Pipeline of the intention-based VQA application.} Past RGB-D, device, and hand trajectories are used to reconstruct the scene and intention prediction. Int3DNet predicts the future intention area, which is projected to 2D image plane and fed into a VLM for reasoning about the intention area.}
  \vspace{-0.3cm}
  \label{fig:fig5}
\end{figure*}

\noindent\textbf{Impact of loss function} We conducted an ablation study to investigate the effect of different loss functions on prediction performance. On the CIRCLE dataset, the Dice loss alone achieved the highest overlap-related metrics (SIM, mIOU, and Dice score). However, its AUC dropped considerably, suggesting limited detection reliability. In contrast, the full loss (BCE, Dice, and Focal) maintained the best AUC while preserving a competitive overlap performance, indicating a balanced trade-off between localization and segmentation quality. On the MoGaze dataset, the Dice loss again yielded a strong overlap performance but suffered from a substantial drop in AUC. Our full loss consistently achieved the best Dice and AUC scores, confirming its robustness in capturing both detection accuracy and spatial overlap. Overall, although the performance differed according to the scene complexity of each dataset, the proposed loss achieved a balanced performance across the metrics. 

\subsection{Demonstration of Intention-based VQA}
To demonstrate the potential applicability of our method to MR scenarios, we further conducted an intention-based VQA pipeline that identifies the object of interaction and retrieves relevant information about it using the ADT dataset~\cite{pan2023aria}. The ADT dataset is a large-scale egocentric dataset captured using Aria smart glasses, featuring 200 sequences of real-world indoor activities with detailed ground-truth data. It includes 6DoF poses of devices and objects, 3D human poses, eye gaze, RGB-D images, and photorealistic synthetic renderings. The dataset provides 3D human pose annotations along with synchronized RGB-D images, making it well-suited for evaluating the proposed method in a simulated MR setting. From the RGB-D images and corresponding camera poses, we reconstructed the scene point clouds to simulate depth perception from the HMD sensors~(\cref{fig:fig5}~(A)). 

\begin{figure}[tb]
  \centering 
  \includegraphics[width=\columnwidth]{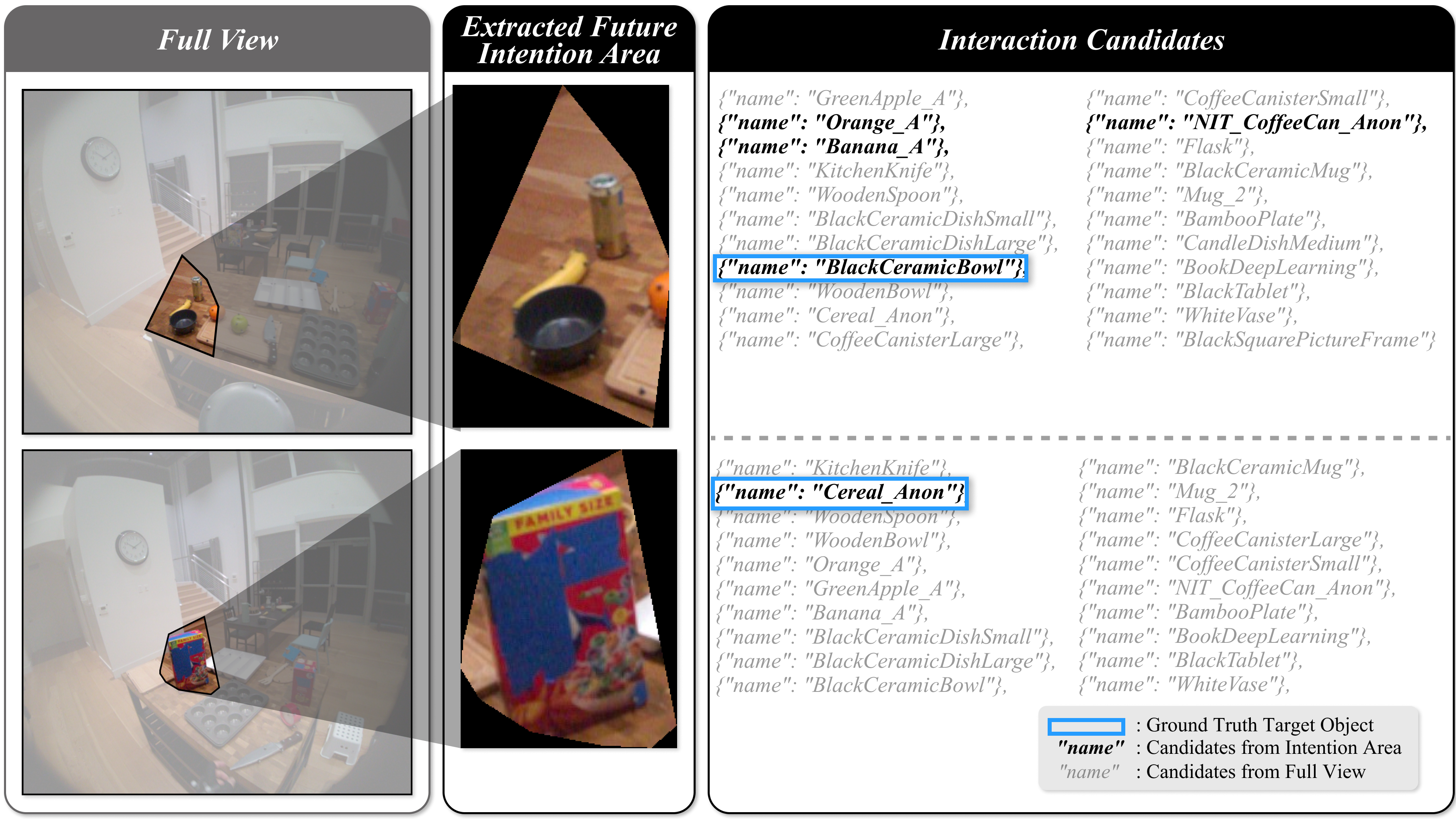}
  \caption{\textbf{Qualitative results of the intention-based VQA.} The right panel shows candidate objects: gray text indicates full-view image outputs, while bold text shows extracted intention area outputs. Blue boxes highlight ground truth target objects. Using the intention area reduces the candidate set for more precise reasoning.}
  \vspace{-0.4cm}
  \label{fig:fig6}
\end{figure}

Leveraging these reconstructions, we implemented intention-based VQA by combining Int3DNet with a foundation vision-language model. Specifically, the reconstructed scene point cloud and the hand–head trajectories are input to the proposed Int3DNet, trained on the CIRCLE dataset, to predict the intention area within the point cloud~(\cref{fig:fig5}~(B)). The predicted 3D intention area is then projected onto the camera coordinates to obtain a 2D area on the image plane. The obtained 2D area was used to crop the corresponding regions of the RGB image~(\cref{fig:fig5}~(C)), which was then provided to the VLM along with a system prompt~(\cref{fig:fig5}~(D)), resulting in a set of candidate objects that the user is likely to interact with. 
\cref{fig:fig6} compares VQA results using the entire image and an image cropped based on user intention. When the entire image is fed into the VLM, the VLM produces a large set of candidate objects (grey text), many of which are irrelevant to the interaction. In contrast, intention-based cropping significantly reduces the number of candidate objects (bold text), including the actual future target object (blue box), by focusing the reasoning on the relevant spatial region. This also reduces visual token usage by approximately 93.6\%, demonstrating the practical benefit of intention prediction for efficient MR reasoning.
Furthermore, as the ADT dataset is unseen during training, the results also demonstrate the generalization ability of our method.

\section{Discussion}
\subsection{Validation of the Proposed Method}

\begin{figure}[tb]
  \centering 
  \includegraphics[width=0.9\columnwidth]{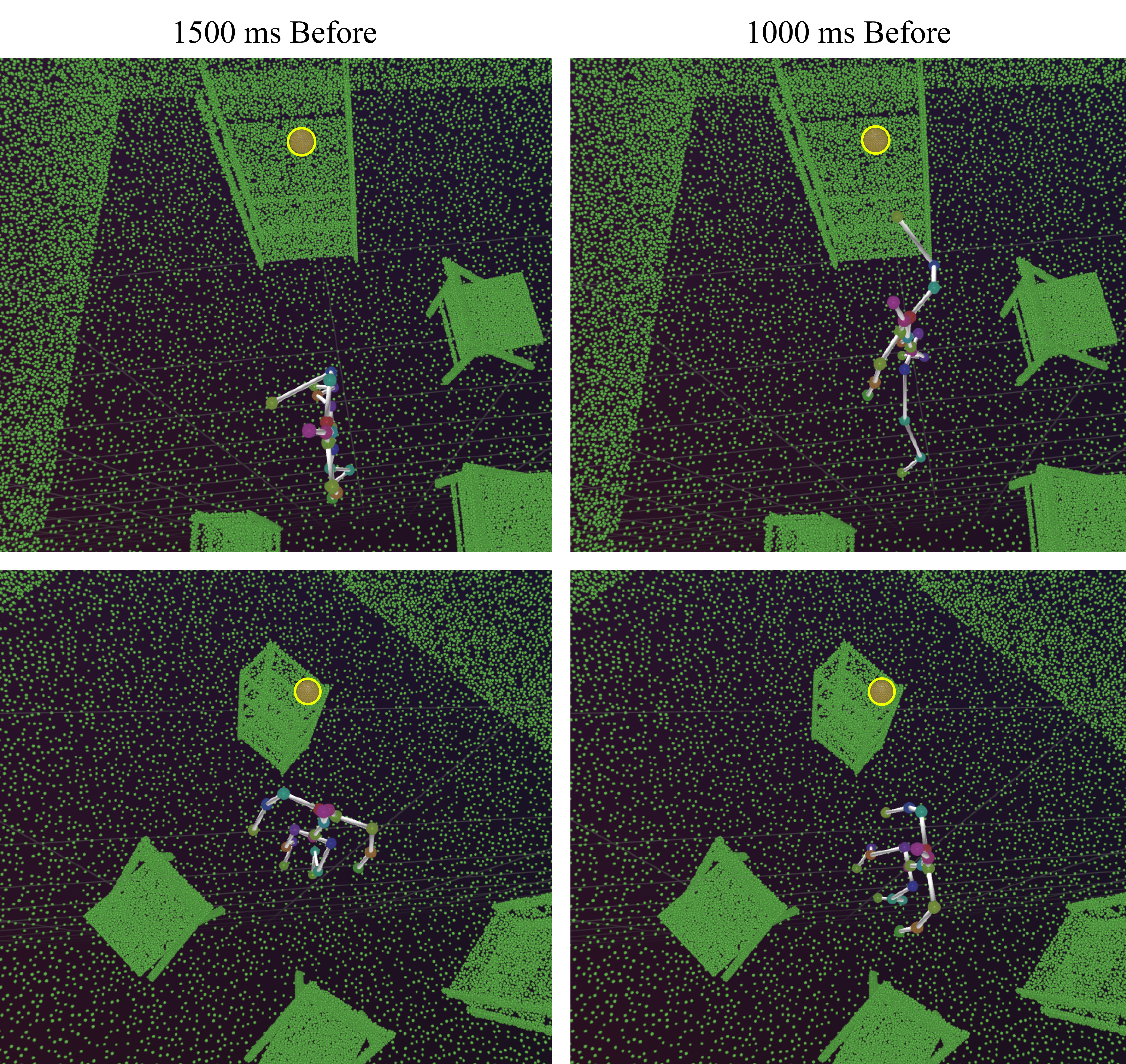}
  \caption{\textbf{Why Performance Drops at 1500 ms in MoGaze.} The MoGaze dataset samples 1500 ms (left) and 1000 ms (right) before the interaction. The yellow sphere represents the target object. 
  }
  \vspace{-0.6cm}
  \label{fig:fig7}
\end{figure}

\label{sec:discussion_1}
The results on the CIRCLE dataset show that Int3DNet achieved consistently better performance than the baselines across metrics and time horizons, indicating reliable performance in complex scenes. On the MoGaze dataset, Int3DNet consistently outperformed the baselines across all metrics until 1200 ms before the interaction. However, beyond this point, the performance advantage diminished, which can be attributed to the characteristics of the dataset. Since the CIRCLE provides a single interaction motion as a segmented sequence, the transitions between interaction targets are clear. As a result, the initiation of an interaction can be recognized as early as 1500 ms before. In contrast, the MoGaze dataset contains continuous motion sequences, including multiple target objects within a single scene. Thus, even at 1500 ms prior to an interaction, the motion often does not yet commit to a target, making the interaction target ambiguous~(\cref{fig:fig7}). While we uniformly segmented motion 1500 ms before each interaction, the uneven lengths of the interaction sequences caused the resulting segments to include motion not intended to interact with the target object. Nevertheless, our Int3DNet maintained high performance until 1200 ms before the interaction, indicating its robustness in intention prediction, even within relatively simple scenes. 

We evaluated the inference latency of Int3DNet to assess its feasibility for MR deployment. On a single NVIDIA GeForce RTX 3090 GPU, the model achieved an average inference time of approximately 113 ms per forward pass, measured over 200 runs after warm-up. While the proposed method is not designed for strict real-time execution, this latency enables near real-time intention prediction for practical MR scenarios. Moreover, Int3DNet is designed for proactive intention prediction, estimating user intention 500–1500 ms before interaction to address the difficulty of reacting precisely at the moment of interaction. The inference latency is well within the target time horizon, allowing downstream processes (\textit{e.g.}, VQA) can be completed prior to actual interaction.

A comparison with the \textit{Head Orientation} model indicates that our method outperforms intention prediction approaches that rely on human attention cues, such as head orientation. This result was consistent across both datasets, regardless of scene complexity, suggesting that integrating scene and motion cues provides more reliable spatial intention inference than relying solely on human attention cues. Moreover, the proposed method outperformed the \textit{Motion Forecast} model. Previously, intention areas within a 3D scene could be inferred by forecasting future motion until the interaction moment. Then, the interaction area was interpreted from the predicted future hand position. However, the results demonstrate that this approach leads to unreliable predictions due to the high degree of freedom in motion forecasting. Overall, the results highlight the advantage of directly leveraging scene geometry for intention prediction compared with approaches that rely solely on motion.

The application demonstration further confirmed that Int3DNet can be effectively applied in real-world MR scenarios. We projected the 3D intention area onto the image plane, enabling the extraction of the user's future interaction region in 2D images. The extracted image is then fed into a VLM model to reason the candidate objects for potential interactions. Compared with feeding the entire image into the VLM model, this approach significantly reduced the number of candidate objects~(\cref{fig:fig6}) while also reducing the number of visual tokens required (approximately 93.6\%). Note that, while we adopted a simple 2D VLM only to demonstrate the usability of intention prediction, adopting 3D scene reasoning could extend the capability to intention areas beyond the FoV and facilitate proactive 3D downstream tasks. Overall, quantitative comparisons with baselines and application demonstration validate both the significance of our method and its usability in real-world applications. 

\begin{table*}[t]
  \caption{\textbf{Attention–Intention Correlation.} Spearman rank correlation coefficient (SRCC) between the attention weights and ground-truth intention area over 15 input motion frames on the CIRCLE and MoGaze datasets. Overall, the results indicate a weak or inconsistent correlation between the learned attention and human intention areas.
  }\label{tab:tab4}
\centering
\resizebox{\textwidth}{!}{
\begin{tabular}{cccccccccccccccccc}
\toprule
  & Frames & 1 & 2 & 3 & 4 & 5 & 6 & 7 & 8 & 9 & 10 & 11 & 12 & 13 & 14 & 15 & Average\\ \midrule
 \multirow{2}{*}{\textit{SRCC}}
  & CIRCLE & -0.0980 & -0.1498 & 0.1470 & -0.1199 & 0.1262 & -0.0571 & 0.0911 & 0.1516 & 0.1741 & -0.0230 & -0.0426 & 0.0700 & -0.0578 & -0.0116 & -0.2434 & -0.0029 \\
  & MoGaze & 0.2654 & -0.1343 & 0.2704 & 0.1024 & 0.2999 & -0.0893 & 0.1434 & 0.2823 & 0.1336 & -0.3595 & -0.3411 & 0.1314 & -0.3785 & -0.2846 & 0.2687 & 0.0207 \\
  \bottomrule  
  \vspace{-1cm}                  
\end{tabular}} 
\end{table*}

\subsection{Effectiveness of Sparse Human Motion Cues}
The fact that consistent performance was achieved even when the test set was composed of unseen scenes indicates Int3DNet's generalization capability, which can reliably predict intention from sparse human motion in an unseen scene. We further evaluated whether our method relied on scene affordances for intention prediction by comparing it with the \textit{Only Scene} model. As shown in \cref{tab:tab1,tab:tab2}, although the \textit{Only Scene} model maintained stable performance across time as it used static scene geometry, its overall accuracy remained lower compared to our method. In the case of the MoGaze dataset, the \textit{Only Scene} model achieved a relatively better performance compared to the CIRCLE dataset. This is likely because the MoGaze dataset contains simple scenes with few static objects (e.g., tables and shelves), allowing the model to roughly estimate the potential interaction areas from the scene geometry alone. This highlights the need for intention prediction research to consider diverse and complex scene configurations. 

Through the comparison with the \textit{Head+Scene} model, we examined whether head orientation could serve as a sufficient cue for predicting the 3D intention area, and thereby evaluated the effectiveness of sparse human motion cues. As shown in \cref{tab:tab1,tab:tab2}, the overall performance was consistently lower than that of sparse human motion cues were incorporated. The results suggest that incorporating human motion with human attention cues significantly improves the representation of human intention. While the \textit{Head+Scene} model outperformed the \textit{Only Scene} model in the CIRCLE dataset, their performance gap was negligible in the MoGaze dataset. This result can also be attributed to scene complexity, where in a simple scene configuration, the scene geometry provides strong cues, thereby the contribution of human attention cues from head orientation becomes negligible. The results also highlight the need of accounting for scene complexity in intention prediction research. 

\subsection{The Relationship Between Attention and Intention}
In neural networks, attention is a mechanism that assigns higher weights to relevant input elements, revealing where the model focuses during computation. This characteristic of the attention mechanism motivated several studies to leverage attention to explain how neural networks work~\cite{lai2020understanding}. While it remains controversial whether attention serves as an explanation~\cite{jain2019attention, wiegreffe2019attention}, investigating the relationship between human intention in the human visual system and the neural attention mechanism is an interesting topic in computer vision research. Several studies~\cite{lai2020understanding, das2017human} systematically experimented the relationship between human attention and neural attention across multiple vision tasks. Their findings suggest that a closer alignment between neural attention and human intention correlates with better performance and interpretability, despite there being certain gaps between neural and human attention. 

Inspired by this, we examined how attention weights relate to the patterns of human intention in 3D scenes. 
We examined the Spearman rank correlation coefficient (SRCC) between the attention map and the ground-truth human intention area to quantitatively assess the relevance of the attention map~(\cref{tab:tab4}). Although we expected a stronger correlation closer to the interaction moment, the overall correlation remained low, with averages of -0.0029 and 0.0207, indicating that attention weights may not serve as explicit indicators of human intention. However, the results of the ablation study show that the cross-attention model has significantly better prediction accuracy than the \textit{MLP-based} model. Moreover, experiments with the \textit{Motion Query} model demonstrated a clear effect of attention directionality, highlighting the role of the attention mechanism in predicting human intention. These results indicate that neural attention mechanisms meaningfully contribute to the prediction of spatial intention regions. In summary, although the experiments yield low SRCC values, the ablation study demonstrates that attention mechanisms significantly improve prediction performance and that attention directionality affects the results. This indicates that attention primarily facilitates task-relevant information flow rather than explicitly representing human intention. Human intention corresponds to high-level behavioral semantics, whereas neural attention emerges as a low-level computational outcome of the optimization process; as a result, attention can influence learning while not necessarily achieving direct alignment with human intention. This is aligned with Lai et al.~\cite{lai2020understanding}, which report that alignment between neural attention and human intention is more meaningful in low-level tasks, while increased task complexity limits such alignment in high-level tasks. Since the current point-level analysis may limit the expression of intention-related patterns, further systematic investigation using object-level prediction is required to clarify the relationship between neural attention and human intention.

\subsection{Limitations and Future Work}
Despite the significance of our method, there are several limitations that we further address in future work. 
First, our study is restricted to pick-and-place interactions. While pick-and-place interactions cover a wide range of hand-based human–scene interactions, full-body and more complex interactions, such as sitting, lying, and tool-use, remain necessary for broader applicability. In future work, we plan to address this limitation by extracting cues from the HMD to infer full-body movements and employing them to capture human-scene interaction intentions, such as where a user is about to sit or lie down. In addition, human intention in tool-use or handover situations may be represented in different wayss. In tool-use interactions, intention can be directed toward both the tool itself and the target object on which the tool is applied. In handover interactions, intention can be directed toward another human body. These diverse human intention representations are challenging to address with the proposed method. Therefore, we plan to explore these domains in further research to extend the applicability of intention prediction.
Moreover, while we demonstrated intention-based VQA on the ADT dataset to show generalization to unseen data and potential use in real-world MR scenarios, the absence of user studies or perceptual evaluations in actual MR environments remains a limitation of the work. We therefore leave user-centered evaluations as important directions for future research, particularly to evaluate how proactive processing through intention prediction affects user experience in real MR deployments.
Finally, as our approach is region-based, multiple objects sometimes appeared within a single predicted intention area when the objects were densely clustered. Even in such cases, the number of target candidates is reduced compared to prediction from the entire view, yet it still remains challenging to identify the exact object. In future work, we plan to adopt a coarse-to-fine strategy, where the proposed method is used to localize a coarse intention area, followed by object-wise fine intention prediction within the narrowed scene using downstream tasks. This approach is expected to enable an efficient use of computational power while accurately predicting the target object.

\section{Conclusion}
In this study, we propose Int3DNet, a cross attention network that fuses head-hand motion with scene geometry for 3D intention prediction in MR. By directly fusing human sparse motion with raw scene point clouds, it is possible to infer spatial intention areas without relying on additional downstream processes. Experiments on the MoGaze and CIRCLE datasets confirmed its robustness across diverse scene complexities and temporal horizons. Furthermore, the integration of Int3DNet into an intention-based VQA pipeline demonstrated its potential for enabling proactive and resource-efficient systems in real-world MR applications. While our study primarily focused on a limited range of interactions, future work will extend intention prediction to full-body, object–object, and human–human interactions that cover diverse scenarios. We believe that Int3DNet will play a vital role in enhancing the efficiency of real-time MR systems, and that our study offers valuable insights into the joint fusion of motion, human attention, and scene geometry for future intention research. 

\acknowledgments{%
	This work was supported by Institute of Information \& communications Technology Planning \& Evaluation (IITP) grant funded by the Korea government (MSIT) (No. RS-2024-00397663, Real-time XR Interface Technology Development for Environmental Adaptation), the National Research Council of Science \& Technology (NST) grant by the Korea government (MSIT) (No. CRC21015), and Korea Institute for Advancement of Technology(KIAT) grant funded by the Korea Government(MOTIE) (RS-2025-02304167, HRD Program for Industrial Innovation). 
}

\bibliographystyle{abbrv-doi-hyperref}
\bibliography{ref}

\begin{thebibliography}{10}

\bibitem{aliakbarian2022flag}
S.~Aliakbarian, P.~Cameron, F.~Bogo, A.~Fitzgibbon, and T.~J. Cashman.
\newblock Flag: Flow-based 3d avatar generation from sparse observations.
\newblock In {\em Proceedings of the IEEE/CVF Conference on Computer Vision and Pattern Recognition}, pp. 13253--13262, 2022.

\bibitem{araujo2023circle}
J.~P. Ara{\'u}jo, J.~Li, K.~Vetrivel, R.~Agarwal, J.~Wu, D.~Gopinath, A.~W. Clegg, and K.~Liu.
\newblock Circle: Capture in rich contextual environments.
\newblock In {\em Proceedings of the IEEE/CVF Conference on Computer Vision and Pattern Recognition}, pp. 21211--21221, 2023.

\bibitem{bao2023uncertainty}
W.~Bao, L.~Chen, L.~Zeng, Z.~Li, Y.~Xu, J.~Yuan, and Y.~Kong.
\newblock Uncertainty-aware state space transformer for egocentric 3d hand trajectory forecasting.
\newblock In {\em Proceedings of the IEEE/CVF international conference on computer vision}, pp. 13702--13711, 2023.

\bibitem{bednarik2012you}
R.~Bednarik, H.~Vrzakova, and M.~Hradis.
\newblock What do you want to do next: a novel approach for intent prediction in gaze-based interaction.
\newblock In {\em Proceedings of the symposium on eye tracking research and applications}, pp. 83--90, 2012.

\bibitem{belardinelli2024gaze}
A.~Belardinelli.
\newblock Gaze-based intention estimation: principles, methodologies, and applications in hri.
\newblock {\em ACM Transactions on Human-Robot Interaction}, 13(3):1--30, 2024.

\bibitem{brunnstrom2020latency}
K.~Brunnstr{\"o}m, E.~Dima, T.~Qureshi, M.~Johanson, M.~Andersson, and M.~Sj{\"o}str{\"o}m.
\newblock Latency impact on quality of experience in a virtual reality simulator for remote control of machines.
\newblock {\em Signal Processing: Image Communication}, 89:116005, 2020.

\bibitem{caserman2019effects}
P.~Caserman, M.~Martinussen, and S.~G{\"o}bel.
\newblock Effects of end-to-end latency on user experience and performance in immersive virtual reality applications.
\newblock In {\em Joint International Conference on Entertainment Computing and Serious Games}, pp. 57--69. Springer, 2019.

\bibitem{chang2021temporal}
Q.~Chang and S.~Zhu.
\newblock Temporal-spatial feature pyramid for video saliency detection.
\newblock {\em arXiv preprint arXiv:2105.04213}, 2021.

\bibitem{chang2023human}
Q.~Chang and S.~Zhu.
\newblock Human vision attention mechanism-inspired temporal-spatial feature pyramid for video saliency detection.
\newblock {\em Cognitive Computation}, 15(3):856--868, 2023.

\bibitem{chen2022gaze}
X.-L. Chen and W.-J. Hou.
\newblock Gaze-based interaction intention recognition in virtual reality.
\newblock {\em Electronics}, 11(10):1647, 2022.

\bibitem{choi2022preemptive}
A.~Choi, M.~K. Jawed, and J.~Joo.
\newblock Preemptive motion planning for human-to-robot indirect placement handovers.
\newblock In {\em 2022 International Conference on Robotics and Automation (ICRA)}, pp. 4743--4749. IEEE, 2022.

\bibitem{clarence2021unscripted}
A.~Clarence, J.~Knibbe, M.~Cordeil, and M.~Wybrow.
\newblock Unscripted retargeting: Reach prediction for haptic retargeting in virtual reality.
\newblock In {\em 2021 IEEE Virtual Reality and 3D User Interfaces (VR)}, pp. 150--159. IEEE, 2021.

\bibitem{das2017human}
A.~Das, H.~Agrawal, L.~Zitnick, D.~Parikh, and D.~Batra.
\newblock Human attention in visual question answering: Do humans and deep networks look at the same regions?
\newblock {\em Computer Vision and Image Understanding}, 163:90--100, 2017.

\bibitem{david2021towards}
B.~David-John, C.~Peacock, T.~Zhang, T.~S. Murdison, H.~Benko, and T.~R. Jonker.
\newblock Towards gaze-based prediction of the intent to interact in virtual reality.
\newblock In {\em ACM symposium on eye tracking research and applications}, pp. 1--7, 2021.

\bibitem{dittadi2021full}
A.~Dittadi, S.~Dziadzio, D.~Cosker, B.~Lundell, T.~J. Cashman, and J.~Shotton.
\newblock Full-body motion from a single head-mounted device: Generating smpl poses from partial observations.
\newblock In {\em Proceedings of the IEEE/CVF International Conference on Computer Vision}, pp. 11687--11697, 2021.

\bibitem{doshi2009roles}
A.~Doshi and M.~M. Trivedi.
\newblock On the roles of eye gaze and head dynamics in predicting driver's intent to change lanes.
\newblock {\em IEEE Transactions on Intelligent Transportation Systems}, 10(3):453--462, 2009.

\bibitem{doshi2012head}
A.~Doshi and M.~M. Trivedi.
\newblock Head and eye gaze dynamics during visual attention shifts in complex environments.
\newblock {\em Journal of vision}, 12(2):9--9, 2012.

\bibitem{du2025predicting}
X.~Du, J.~Wu, X.~Tang, X.~Lv, L.~Jia, and C.~Xue.
\newblock Predicting user attention states from multimodal eye--hand data in vr selection tasks.
\newblock {\em Electronics}, 14(10):2052, 2025.

\bibitem{frischen2007gaze}
A.~Frischen, A.~P. Bayliss, and S.~P. Tipper.
\newblock Gaze cueing of attention: visual attention, social cognition, and individual differences.
\newblock {\em Psychological bulletin}, 133(4):694, 2007.

\bibitem{gamage2021so}
N.~M. Gamage, D.~Ishtaweera, M.~Weigel, and A.~Withana.
\newblock So predictable! continuous 3d hand trajectory prediction in virtual reality.
\newblock In {\em The 34th Annual ACM Symposium on User Interface Software and Technology}, pp. 332--343, 2021.

\bibitem{goldstein2002sensation}
E.~B. Goldstein and J.~R. Brockmole.
\newblock {\em Sensation and perception}, vol.~90.
\newblock Wadsworth-Thomson Learning Pacific Grove, CA, 2002.

\bibitem{hayes2021deep}
T.~R. Hayes and J.~M. Henderson.
\newblock Deep saliency models learn low-, mid-, and high-level features to predict scene attention.
\newblock {\em Scientific reports}, 11(1):18434, 2021.

\bibitem{he2018real}
H.~He, Y.~She, J.~Xiahou, J.~Yao, J.~Li, Q.~Hong, and Y.~Ji.
\newblock Real-time eye-gaze based interaction for human intention prediction and emotion analysis.
\newblock In {\em Proceedings of computer graphics international 2018}, pp. 185--194. 2018.

\bibitem{hoffman2023inferring}
G.~Hoffman, T.~Bhattacharjee, and S.~Nikolaidis.
\newblock Inferring human intent and predicting human action in human--robot collaboration.
\newblock {\em Annual Review of Control, Robotics, and Autonomous Systems}, 7, 2023.

\bibitem{hu2022we}
Z.~Hu, D.~Yang, S.~Cheng, L.~Zhou, S.~Wu, and J.~Liu.
\newblock We know where they are looking at from the rgb-d camera: Gaze following in 3d.
\newblock {\em IEEE Transactions on Instrumentation and Measurement}, 71:1--14, 2022.

\bibitem{hu2024hoimotion}
Z.~Hu, Z.~Yin, D.~Haeufle, S.~Schmitt, and A.~Bulling.
\newblock Hoimotion: Forecasting human motion during human-object interactions using egocentric 3d object bounding boxes.
\newblock {\em IEEE Transactions on Visualization and Computer Graphics}, 2024.

\bibitem{huang2025human}
Y.~Huang, D.~Fan, D.~Yan, W.~Qi, G.~Deng, Z.~Shao, Y.~Luo, D.~Li, Z.~Wang, Q.~Liu, et~al.
\newblock Human-robot collaborative tele-grasping in clutter with five-fingered robotic hands.
\newblock {\em IEEE Robotics and Automation Letters}, 2025.

\bibitem{jain2019attention}
S.~Jain and B.~C. Wallace.
\newblock Attention is not explanation.
\newblock {\em arXiv preprint arXiv:1902.10186}, 2019.

\bibitem{jiang2022avatarposer}
J.~Jiang, P.~Streli, H.~Qiu, A.~Fender, L.~Laich, P.~Snape, and C.~Holz.
\newblock Avatarposer: Articulated full-body pose tracking from sparse motion sensing.
\newblock In {\em European conference on computer vision}, pp. 443--460. Springer, 2022.

\bibitem{katharopoulos2020transformers}
A.~Katharopoulos, A.~Vyas, N.~Pappas, and F.~Fleuret.
\newblock Transformers are rnns: Fast autoregressive transformers with linear attention.
\newblock In {\em International conference on machine learning}, pp. 5156--5165. PMLR, 2020.

\bibitem{kedia2024interact}
K.~Kedia, A.~Bhardwaj, P.~Dan, and S.~Choudhury.
\newblock Interact: Transformer models for human intent prediction conditioned on robot actions.
\newblock In {\em 2024 IEEE International Conference on Robotics and Automation (ICRA)}, pp. 621--628. IEEE, 2024.

\bibitem{koochaki2019eye}
F.~Koochaki and L.~Najafizadeh.
\newblock Eye gaze-based early intent prediction utilizing cnn-lstm.
\newblock In {\em 2019 41st Annual international conference of the IEEE engineering in medicine and biology society (EMBC)}, pp. 1310--1313. IEEE, 2019.

\bibitem{kratzer2020mogaze}
P.~Kratzer, S.~Bihlmaier, N.~B. Midlagajni, R.~Prakash, M.~Toussaint, and J.~Mainprice.
\newblock Mogaze: A dataset of full-body motions that includes workspace geometry and eye-gaze.
\newblock {\em IEEE Robotics and Automation Letters}, 6(2):367--373, 2020.

\bibitem{kratzer2020anticipating}
P.~Kratzer, N.~B. Midlagajni, M.~Toussaint, and J.~Mainprice.
\newblock Anticipating human intention for full-body motion prediction in object grasping and placing tasks.
\newblock In {\em 2020 29th IEEE international conference on robot and human interactive communication (RO-MAN)}, pp. 1157--1163. IEEE, 2020.

\bibitem{lai2020understanding}
Q.~Lai, S.~Khan, Y.~Nie, H.~Sun, J.~Shen, and L.~Shao.
\newblock Understanding more about human and machine attention in deep neural networks.
\newblock {\em IEEE Transactions on Multimedia}, 23:2086--2099, 2020.

\bibitem{land1999roles}
M.~Land, N.~Mennie, and J.~Rusted.
\newblock The roles of vision and eye movements in the control of activities of daily living.
\newblock {\em Perception}, 28(11):1311--1328, 1999.

\bibitem{laplaza2022context}
J.~Laplaza, A.~Garrell, F.~Moreno-Noguer, and A.~Sanfeliu.
\newblock Context and intention for 3d human motion prediction: experimentation and user study in handover tasks.
\newblock In {\em 2022 31st IEEE international conference on robot and human interactive communication (RO-MAN)}, pp. 630--635. IEEE, 2022.

\bibitem{laplaza2024enhancing}
J.~Laplaza, F.~Moreno, and A.~Sanfeliu.
\newblock Enhancing robotic collaborative tasks through contextual human motion prediction and intention inference.
\newblock {\em International Journal of Social Robotics}, pp. 1--20, 2024.

\bibitem{lee2023questenvsim}
S.~Lee, S.~Starke, Y.~Ye, J.~Won, and A.~Winkler.
\newblock Questenvsim: Environment-aware simulated motion tracking from sparse sensors.
\newblock In {\em ACM SIGGRAPH 2023 Conference Proceedings}, pp. 1--9, 2023.

\bibitem{leon2022intuitive}
J.~F. Leon, D.~Gonzalez-Aguirre, and L.~Nachman.
\newblock Intuitive and efficient human-robot collaboration via real-time approximate bayesian inference.
\newblock {\em arXiv preprint arXiv:2205.08657}, 2022.

\bibitem{lin2017focal}
T.-Y. Lin, P.~Goyal, R.~Girshick, K.~He, and P.~Doll{\'a}r.
\newblock Focal loss for dense object detection.
\newblock In {\em Proceedings of the IEEE international conference on computer vision}, pp. 2980--2988, 2017.

\bibitem{liu2021motion}
Z.~Liu, P.~Su, S.~Wu, X.~Shen, H.~Chen, Y.~Hao, and M.~Wang.
\newblock Motion prediction using trajectory cues.
\newblock In {\em Proceedings of the IEEE/CVF international conference on computer vision}, pp. 13299--13308, 2021.

\bibitem{luo2019human}
R.~C. Luo and L.~Mai.
\newblock Human intention inference and on-line human hand motion prediction for human-robot collaboration.
\newblock In {\em 2019 IEEE/RSJ International Conference on Intelligent Robots and Systems (IROS)}, pp. 5958--5964. IEEE, 2019.

\bibitem{ma2022progressively}
T.~Ma, Y.~Nie, C.~Long, Q.~Zhang, and G.~Li.
\newblock Progressively generating better initial guesses towards next stages for high-quality human motion prediction.
\newblock In {\em Proceedings of the IEEE/CVF conference on computer vision and pattern recognition}, pp. 6437--6446, 2022.

\bibitem{mainprice2013human}
J.~Mainprice and D.~Berenson.
\newblock Human-robot collaborative manipulation planning using early prediction of human motion.
\newblock In {\em 2013 IEEE/RSJ International Conference on Intelligent Robots and Systems}, pp. 299--306. IEEE, 2013.

\bibitem{milletari2016v}
F.~Milletari, N.~Navab, and S.-A. Ahmadi.
\newblock V-net: Fully convolutional neural networks for volumetric medical image segmentation.
\newblock In {\em 2016 fourth international conference on 3D vision (3DV)}, pp. 565--571. Ieee, 2016.

\bibitem{nguyen2023open}
T.~Nguyen, M.~N. Vu, A.~Vuong, D.~Nguyen, T.~Vo, N.~Le, and A.~Nguyen.
\newblock Open-vocabulary affordance detection in 3d point clouds.
\newblock In {\em 2023 IEEE/RSJ International Conference on Intelligent Robots and Systems (IROS)}, pp. 5692--5698. IEEE, 2023.

\bibitem{pan2023aria}
X.~Pan, N.~Charron, Y.~Yang, S.~Peters, T.~Whelan, C.~Kong, O.~Parkhi, R.~Newcombe, and Y.~C. Ren.
\newblock Aria digital twin: A new benchmark dataset for egocentric 3d machine perception.
\newblock In {\em Proceedings of the IEEE/CVF International Conference on Computer Vision}, pp. 20133--20143, 2023.

\bibitem{pettersson2024intended}
J.~Pettersson and P.~Falkman.
\newblock Intended human arm movement direction prediction using eye tracking.
\newblock {\em International Journal of Computer Integrated Manufacturing}, 37(9):1107--1125, 2024.

\bibitem{ponton2023sparseposer}
J.~L. Ponton, H.~Yun, A.~Aristidou, C.~Andujar, and N.~Pelechano.
\newblock Sparseposer: Real-time full-body motion reconstruction from sparse data.
\newblock {\em ACM Transactions on Graphics}, 43(1):1--14, 2023.

\bibitem{qi2017pointnet++}
C.~R. Qi, L.~Yi, H.~Su, and L.~J. Guibas.
\newblock Pointnet++: Deep hierarchical feature learning on point sets in a metric space.
\newblock {\em Advances in neural information processing systems}, 30, 2017.

\bibitem{shao2025great}
Y.~Shao, W.~Zhai, Y.~Yang, H.~Luo, Y.~Cao, and Z.-J. Zha.
\newblock Great: Geometry-intention collaborative inference for open-vocabulary 3d object affordance grounding.
\newblock In {\em Proceedings of the Computer Vision and Pattern Recognition Conference}, pp. 17326--17336, 2025.

\bibitem{shi2021understand}
X.~Shi, Y.~Yang, and Q.~Liu.
\newblock I understand you: Blind 3d human attention inference from the perspective of third-person.
\newblock {\em IEEE Transactions on Image Processing}, 30:6212--6225, 2021.

\bibitem{wang2025grace}
C.~Wang, W.~Zhai, Y.~Yang, Y.~Cao, and Z.~Zha.
\newblock Grace: Estimating geometry-level 3d human-scene contact from 2d images.
\newblock {\em arXiv preprint arXiv:2505.06575}, 2025.

\bibitem{wei2017inferring}
P.~Wei, D.~Xie, N.~Zheng, S.-C. Zhu, et~al.
\newblock Inferring human attention by learning latent intentions.
\newblock In {\em IJCAI}, pp. 1297--1303, 2017.

\bibitem{wiegreffe2019attention}
S.~Wiegreffe and Y.~Pinter.
\newblock Attention is not not explanation.
\newblock {\em arXiv preprint arXiv:1908.04626}, 2019.

\bibitem{winkler2022questsim}
A.~Winkler, J.~Won, and Y.~Ye.
\newblock Questsim: Human motion tracking from sparse sensors with simulated avatars.
\newblock In {\em SIGGRAPH Asia 2022 Conference Papers}, pp. 1--8, 2022.

\bibitem{yan2023gazemodiff}
H.~Yan, Z.~Hu, S.~Schmitt, and A.~Bulling.
\newblock Gazemodiff: Gaze-guided diffusion model for stochastic human motion prediction.
\newblock {\em arXiv preprint arXiv:2312.12090}, 2023.

\bibitem{yang2023learning}
S.~Yang, N.~P. Garg, R.~Gao, M.~Yuan, B.~Noronha, W.~T. Ang, and D.~Accoto.
\newblock Learning-based motion-intention prediction for end-point control of upper-limb-assistive robots.
\newblock {\em Sensors}, 23(6):2998, 2023.

\bibitem{yeamkuan20213d}
S.~Yeamkuan, K.~Chamnongthai, and W.~Pichitwong.
\newblock A 3d point-of-intention estimation method using multimodal fusion of hand pointing, eye gaze and depth sensing for collaborative robots.
\newblock {\em IEEE Sensors Journal}, 22(3):2700--2710, 2021.

\bibitem{yin2024robust}
Z.~Yin, Z.~Wan, M.~Yang, Y.~Xiong, W.~Wang, and S.~Wu.
\newblock Robust gaze-based intention prediction for real-world scenarios.
\newblock {\em IEEE Transactions on Cognitive and Developmental Systems}, 2024.

\bibitem{yu2015human}
Z.~Yu and M.~Lee.
\newblock Human motion based intent recognition using a deep dynamic neural model.
\newblock {\em Robotics and Autonomous Systems}, 71:134--149, 2015.

\bibitem{zhao2020research}
M.~Zhao, H.~Gao, W.~Wang, and J.~Qu.
\newblock Research on human-computer interaction intention recognition based on eeg and eye movement.
\newblock {\em IEEE Access}, 8:145824--145832, 2020.

\end{thebibliography}
\end{document}